\documentclass{ieeeaccess}
\usepackage{cite}
\usepackage{amsmath,amssymb,amsfonts}
\usepackage{multirow}
\newcommand*\rot{\rotatebox{90}}

\usepackage{algorithm}
\usepackage{algpseudocode}

\usepackage{graphicx}
\usepackage{textcomp}
\usepackage{hyperref}
\usepackage{color,soul}
\usepackage{tabularx}
\usepackage{ragged2e}
\def\BibTeX{{\rm B\kern-.05em{\sc i\kern-.025em b}\kern-.08em
    T\kern-.1667em\lower.7ex\hbox{E}\kern-.125emX}}
\begin{document}
\history{Date of publication xxxx 00, 0000, date of current version xxxx 00, 0000.}
\doi{10.1109/ACCESS}

\title{A Novel Transformer-based Self-Supervised Learning Method to Enhance Photoplethysmogram Signal Artifact Detection}
\author{\uppercase{Thanh-Dung Le}\authorrefmark{1,2},  \IEEEmembership{Member, IEEE}, \uppercase{Clara Macabiau}\authorrefmark{1}, \uppercase{K\'evin Albert}\authorrefmark{3}, \uppercase{Philippe Jouvet}\authorrefmark{3}, \uppercase{and Rita Noumeir}\authorrefmark{1},
\IEEEmembership{Member, IEEE}}

\address[1]{Biomedical Information Processing Lab, \'{E}cole de Technologie Sup\'{e}rieure, Montr\'{e}al, Qu\'{e}bec, Canada}

\address[2]{The Interdisciplinary Centre for Security, Reliability, and Trust (SnT), University of Luxembourg, Luxembourg}
\address[3]{CHU Sainte-Justine Research Center, CHU Sainte-Justine Hospital, University of Montreal, Montr\'{e}al, Qu\'{e}bec, Canada}

\tfootnote{This work was supported in part by the Natural Sciences and Engineering Research Council (NSERC), in part by the Institut de Valorisation des données de l’Université de Montréal (IVADO), in part by the Fonds de la recherche en sante du Quebec (FRQS).}

\markboth
{Thanh-Dung Le \headeretal: A Novel Transformer-based Self-Supervised Learning Method to Enhance Photoplethysmogram Signal Artifact Detection}
{Thanh-Dung Le \headeretal: A Novel Transformer-based Self-Supervised Learning Method to Enhance Photoplethysmogram Signal Artifact Detection}

\corresp{Corresponding author: Thanh-Dung Le (e-mail: thanh-dung.le@uni.lu).}

\begin{abstract} 
Recent research has revealed that traditional machine learning methods, such as semi-supervised label propagation and K-nearest neighbors, outperform Transformer-based models in artifact detection from photoplethysmogram (PPG) signals, mainly when data is limited. This study addresses the underutilization of abundant unlabeled data by employing self-supervised learning (SSL) to extract latent features from these data, followed by fine-tuning on labeled data. Our experiments demonstrate that SSL significantly enhances the Transformer model's ability to learn representations, improving its robustness in artifact classification tasks. Among various SSL techniques—including masking, contrastive learning, and DINO (self-distillation with no labels)—contrastive learning exhibited the most stable and superior performance in small PPG datasets. Further, we delve into optimizing contrastive loss functions, which are crucial for contrastive SSL. Inspired by InfoNCE, we introduce a novel contrastive loss function that facilitates smoother training and better convergence, thereby enhancing performance in artifact classification. In summary, this study establishes the efficacy of SSL in leveraging unlabeled data, particularly in enhancing the capabilities of the Transformer model in PPG artifact detection. This approach holds promise for broader applications in PICU environments, where annotated data is often limited.
\end{abstract}


\titlepgskip=-21pt

\maketitle

\section{Introduction}
\label{sec:introduction}
\PARstart{R}{ecently} the Pediatric Intensive Care Unit (PICU) at CHU Sainte-Justine (CHUSJ) has made notable advancements by developing a high-resolution research database (HRDB) \cite{brossier2018creating, roumeliotis2018reorganizing}. This innovative database directly integrates biomedical signals from various monitoring devices into the electronic patient record, ensuring seamless data incorporation throughout a patient's PICU stay \cite{mathieu2021validation}. The integration of HRDB has significantly enhanced the Clinical Decision Support System (CDSS) at CHUSJ, boosting patient safety and underpinning decision-making with robust evidence \cite{dziorny2022clinical}. In this context, early and accurate diagnosis of acute respiratory distress syndrome (ARDS) is a pivotal goal of the CDSS at CHUSJ. Oxygen saturation (SpO\textsubscript{2}) values, critical in ARDS diagnosis, are central to predicting and managing ARDS \cite{le2022detecting, sauthier2021estimated}, and play a key role in determining respiratory support strategies \cite{emeriaud2023executive, jouvet2012pilot, wysocki2014closed}. To have a valid SpO\textsubscript{2} value, it is necessary to have a good signal, i.e., Photoplesthymography waveform. Moreover, the PPG waveforms can also be used in algorithms to estimate non-invasive blood pressure \cite{hill2021imputation, fan2017estimating}. Consequently, accurately identifying and removing erroneous waveforms and SpO\textsubscript{2} values from CDSS inputs is crucial. Ensuring the reliability of these inputs is essential for the effective functioning of the CDSS, directly impacting patient outcomes and care efficiency.

Building on the foundation of our work in fully-supervised and semi-supervised learning methodologies, we have delved into PPG artifact detection, focusing on machine learning (ML) applications. A pivotal study in this area \cite{macabiau2023label} investigated the use of machine learning techniques for this purpose. However, challenges arise in scenarios featuring imbalanced classes and limited data availability. In these contexts, Transformer models, despite their advanced attention mechanisms, have exhibited suboptimal performance compared to other methods, such as semi-supervised label propagation and supervised KNN learning. The core issue lies in the Transformer models' reduced efficacy in smaller datasets. To address these limitations and enhance the Transformer's applicability in artifact detection, our recent study \cite{le2023grn} introduced an innovative approach. We incorporated the Gated Residual Network (GRN) into the Transformer framework, enhancing its performance capabilities significantly. This GRN-Transformer hybrid model not only overcomes the inherent limitations of traditional Transformer models in handling smaller datasets but also outperforms other existing models in artifact detection accuracy and reliability.
Despite these advancements in artifact detection, a standard limitation persists across recent studies: a heavy reliance on annotated data to train fully supervised machine learning algorithms. This reliance is particularly evident considering that in most cases, only up to 10\% of the data is annotated, leaving a vast 90\% of the data pool unexploited. Such underutilization of available data presents a significant challenge, especially when dealing with small datasets and imbalanced classes, which are common in binary classification for motion artifact detection in PPG signals.  In light of these challenges, this study aims to transcend the constraints of labeled data dependency by harnessing the potential of SSL. This approach is particularly pertinent for two key reasons as follows: i) The vast majority of our data, approximately 90\%, remain unannotated, representing a largely untapped resource that could significantly enhance our understanding and detection capabilities, and ii) There is a compelling opportunity to explore how SSL can adapt and perform with limited and unlabeled data, a scenario frequently encountered in clinical settings. By pivoting towards SSL, we aim to leverage the underused unannotated data, potentially revolutionizing how we approach artifact detection and enhancing the robustness of classifiers in these challenging PICU environments.

\begin{figure*}[!t]
	\centering
	\includegraphics[scale=0.275]{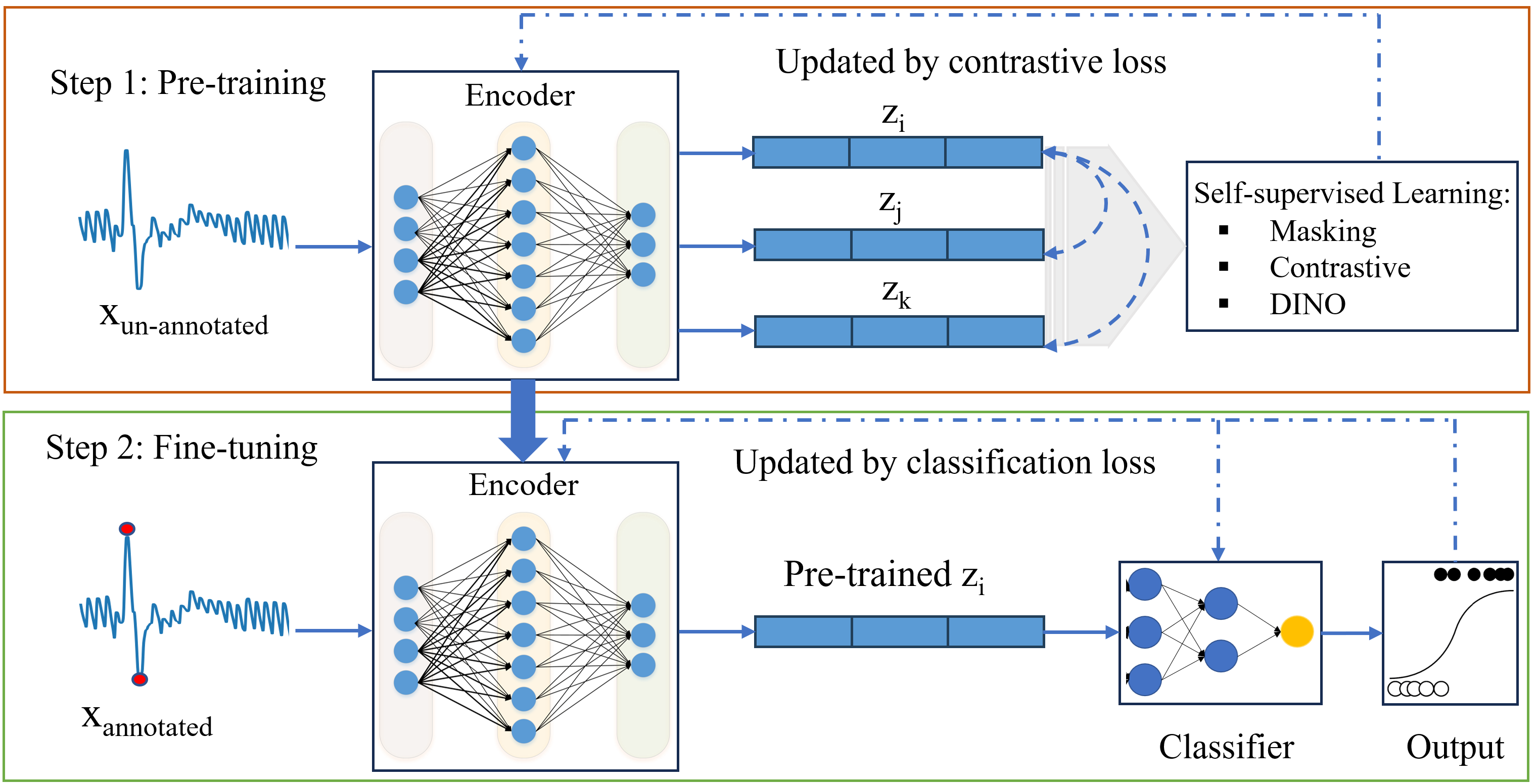}
	\caption{An end-end process diagram workflow demonstration for the proposed two-stage framework for PPG signal analysis. Step 1 involves pre-training the encoder on unannotated data using different self-supervised learning techniques (masking, contrastive, and DINO) to learn optimal hidden representations. Step 2 fine-tunes the pre-trained encoder with a classifier using annotated data to optimize the classification loss, thereby improving model performance with minimal labeled data.}
	\label{fig:workflow}
\end{figure*}

In exploring SSL's effectiveness for PPG artifact classification, as depicted in Fig. \ref{fig:workflow}, we implemented three distinct SSL strategies: masking, contrastive learning, and self-distillation without labels (DINO). Masking involves concealing parts of the input data and training the model to reconstruct these hidden segments, enhancing its capability to capture local and global data features. Contrastive learning, on the other hand, optimizes the model by learning to distinguish between similar and dissimilar data points, refining the representation space to capture subtle patterns in the signal effectively. DINO employs a teacher-student architecture, where the student model mimics the teacher's output, even without labeled data, promoting the extraction of meaningful representations.

The framework is divided into two stages. The first stage focuses on pre-training the encoder network using various SSL techniques, such as masking, contrastive learning, and DINO. This step aims to learn optimal hidden representations, denoted as $z_i$, for the unannotated training data $X_{un-annotated}$. This is achieved by minimizing a contrastive loss specific to each SSL technique, which encourages the encoder to generate meaningful embeddings that capture the intrinsic structure of the data. These embeddings $z_i, z_j$, and $z_k$ represent different transformations of the same signal or similar signals, which are contrasted against each other to refine the representation. Second stage, the learned hidden representations from the pre-trained encoder $z_i$ are transferred to a classifier network. This step involves fine-tuning the encoder along with the classifier using a small set of annotated data $X_{annotated}$. The goal is to optimize the classification loss, which measures the discrepancy between the predicted labels from the classifier and the true annotated labels. By leveraging the pre-trained representations, the framework improves performance on the classification task with minimal labeled data.

The combined approach ensures the effective utilization of unannotated data during the pre-training stage and maximizes the impact of the limited annotated data during the fine-tuning stage. The experimental results revealed that this combination of pre-training on unlabeled data and fine-tuning on just 10\% of annotated data substantially enhanced classification performance compared to models trained using traditional supervised learning on limited data. Contrastive learning demonstrated the highest efficacy among the three SSL methods, underscoring the critical role of contrastive loss function design. To optimize this, we refined the standard InfoNCE loss, leading to a more stable training process and better convergence, significantly boosting the model’s artifact detection performance. Although unsupervised learning techniques contributed positively, they could not match the performance gains achieved with SSL, emphasizing the superiority of leveraging structured pre-training strategies for improving model robustness and generalization.

\section{Related Works}

The accuracy and reliability of pulse oximeters, essential for monitoring blood oxygen levels, are significantly impacted by motion artifacts in PPG signals \cite{khan2015robust}. Traditional filtering algorithms, while helpful, are not fully effective in eliminating these motion artifacts. As a result, residual motion artifacts often lead to inaccuracies in measuring blood oxygen saturation. This underscores the importance of developing more sophisticated methods for detecting and mitigating motion artifacts in PPG signals to ensure the reliability and precision of pulse oximetry readings. In the PICU setting, the necessity for accurate PPG artifact detection becomes even more pronounced \cite{macabiau2023label}. Children, especially those in critical care, are often prone to more frequent and abrupt movements than adults. Therefore, enhancing PPG signal processing to identify and compensate for these artifacts accurately is vital for improving patient monitoring and care quality in pediatric intensive care settings.

The integration of ML into PPG analysis has revolutionized various clinical applications, addressing diverse healthcare needs with greater precision. A notable example is heart rate estimation, where ML models have shown exceptional accuracy, as highlighted in the studies by Dao et al. \cite{dao2016robust}, and Mehrgardt et al. \cite{mehrgardt2021deep}, with accuracies greater than 95\%. This advancement extends beyond basic applications; ML also enables real-time physiological monitoring by providing up to over 18\% improvements in prediction accuracy. For example, it provides critical insights into vital signs such as oxygen saturation \cite{fan2017estimating}, blood pressure \cite{liu2020pca}, and respiratory rate \cite{birrenkott2017robust}, thereby enhancing patient care and enabling the early identification of potential health issues, as shown in research by Venema et al. \cite{venema2013robustness} and Alharbi et al. \cite{alharbi2022non}, achieving over 80\% for precision, and recall for prediction of patient condition. Furthermore, ML algorithms have been particularly effective in detecting and filtrating motion artifacts in PPG data, a crucial step in ensuring the reliability of continuous monitoring systems, as demonstrated by Nwibor et al. \cite{nwibor2023remote} with 98.7\% accuracy. Consequently, the application of ML in PPG analysis has not only broadened the scope of clinical applications but has also significantly increased the efficiency and accuracy of healthcare interventions.

The implementation of conventional ML techniques marked initial advancements in this field. For instance, Support Vector Machine classifiers have effectively detected heart rates from PPG signals, employing time-frequency spectral features \cite{dao2016robust}, with accuracies greater than 95\%. This approach represents the foundational phase of machine learning applications in PPG analysis. However, the field has witnessed a significant shift towards more sophisticated methods, particularly with the advent of deep learning algorithms. Studies have increasingly adopted deep learning models like Multilayer Perceptrons and Fully Convolutional Neural Networks, which have shown promising results in artifact detection \cite{wang2017time, marzorati2022hybrid}, accuracy greater than 90\%. These advancements reflect a transition from conventional machine learning techniques to more complex and capable deep learning methods within the supervised learning paradigm, significantly enhancing the capabilities in PPG signal analysis.

Recent advancements in PPG signal analysis underscore a paradigm shift from traditional supervised learning methods to exploring the potential of unsupervised and semi-supervised approaches. Initial studies, such as the one by Maqsood et al. \cite{maqsood2021benchmark}, have demonstrated the effectiveness of deep learning algorithms, particularly Bi-LSTM combined with time-domain features, in achieving superior heart rate estimation. This approach primarily relied on supervised learning frameworks. However, our research team's study \cite{macabiau2023label} marks a significant transition in this domain, achieving a precision of 91\%, a recall of 90\%, and an F1 score of 90\%. We explored a range of ML techniques, encompassing not only conventional machine learning and supervised models like MLP and Transformer but also semi-supervised learning with label propagation. Interestingly, we found that semi-supervised label propagation and the supervised KNN algorithm exhibited better performance than Transformer models, especially in scenarios involving imbalanced classes and limited data. This finding indicates a growing trend toward leveraging semi-supervised methods in PPG artifact detection, highlighting their potential to overcome data scarcity and class imbalance challenges.

In light of the demonstrated advantages of SSL, this study will focus primarily on this approach, particularly its effectiveness in providing robust representations for downstream tasks without the need for labeled data.  SSL can significantly enhance robustness in various aspects, including resistance to adversarial examples, resilience against label corruption, and tolerance to common input corruption. Moreover, an intriguing aspect of self-supervision is its remarkable capability to aid out-of-distribution detection, especially with challenging near-distribution outliers. In fact, in these scenarios, self-supervision has been observed to surpass the performance of fully supervised methods, as detailed in the study by Hendrycks et al. \cite{hendrycks2019using}. These insights demonstrate the potential of self-supervision in improving robustness and uncertainty estimation and establish these domains as critical avenues for future research in SSL. Our study, therefore, seeks to delve deeper into these promising aspects of self-supervision, aiming to contribute meaningful advancements in artifact detection.

In conclusion, the integration of ML in PPG analysis has marked a significant milestone in the field, especially in enhancing artifact detection. This advancement has improved signal reliability and opened new frontiers in clinical applications. However, this domain still requires extensive research and development to fine-tune these methodologies and ensure their effective integration into clinical practices. This ongoing progression is set to redefine the landscape of PPG signal analysis and its consequential impact on patient care, particularly in the PICU at CHUSJ. Additionally, despite the transformative potential of Transformer models and attention mechanisms in ML, as surveyed by Lin et al. \cite{lin2022survey}, their efficacy in small dataset scenarios remains a challenge, as highlighted by our recent studies \cite{macabiau2023label, mehrgardt2021deep, le2023small}. This study, therefore, was driven by the objective of augmenting the SSL capability in effectively managing small datasets and imbalanced classes. Our focus is particularly on the classification task of detecting motion artifacts in PPG signals, a critical aspect of ensuring accurate and reliable patient monitoring.

\section{Materials and Methods}
\subsection{Clinical PPG Data at CHUSJ}
\label{sec:data_CHUSJ}

The PICU at CHUSJ has implemented a high-resolution research database (HRDB), which has received approval from the ethical committee. This HRDB is a comprehensive system that effectively integrates biomedical signals from various monitoring devices with electronic patient records. This integration occurs throughout each patient's hospital stay, providing a rich and detailed dataset for research purposes.

The research protocol conducted for this study received approval from the research ethics board of CHUSJ, University of Montreal, under the project number eNIMP:2023-4556. The data collection process within this HRDB captures a wide range of physiological signals. Key among these was the pulse oximeter sensor, which played a crucial role in acquiring PPG signals. This sensor works by emitting light into the skin and measuring the variations in light absorption due to blood flow changes during the cardiac cycle. 

The study's population included children aged from newborn to 18 years who were admitted to CHUSJ between September 2018 and September 2023. The inclusion criteria were based on the availability of essential waveform records such as PPG,  electrocardiogram (ECG), and arterial brood pressure (ABP). Certain exclusion criteria were applied to maintain the integrity and quality of the data. Any data collected after the fourth day of hospitalization due to prolonged PICU stay were excluded to minimize potential biases. Moreover, the study did not include patients undergoing Extracorporeal Membrane Oxygenation (ECMO) treatment. In cases of multiple readmissions, only data from the initial admission were considered to maintain data independence and avoid confounding factors.

The final cohort comprised 1,571 eligible patients. For each patient, continuous recordings of ECG, PPG, and blood pressure via catheter were made over a 96-hour period. The PPG signals were specifically captured every 5 seconds at a sampling frequency of 128 Hz. Similarly, blood pressure and ECG signals were acquired with a frequency of 512 Hz, also at 5-second intervals. During the data extraction phase, a fixed 30-second window was employed for each PPG signal to facilitate subsequent processing and analysis.

\subsection{Data Pre-Processing}
\label{sec:data_prep}
In our study, data preprocessing plays a crucial role in enhancing the quality of PPG signals, adhering to the methodologies established in our prior studies \cite{macabiau2023label, le2023grn}. The raw PPG signal undergoes a series of preprocessing steps to enhance signal quality, eliminate noise, and prepare it for subsequent analysis stages. This process encompasses four main steps: filtering, segmentation, resampling, and normalization, detailed as follows:
\begin{enumerate}
\item \textbf{Filtering}: Each signal segment is processed using a band-pass Butterworth filter with cut-off frequencies set to 0.5 Hz and 5 Hz, corresponding to heart rates ranging from 30 to 300 beats per minute. A forward-backward filtering approach is applied to preserve the phase integrity of the signal. This step helps eliminate baseline drifts and high-frequency noise components.
\item \textbf{Pulse segmentation}: Local minima detection is employed to segment the signal, defining each pulse between consecutive minima. This segmentation strategy isolates individual pulses, making analyzing and detecting artifacts within each segment easier. The dynamic segment size varies based on signal characteristics and the desired analysis objectives.
\item \textbf{Resampling}: Given that a cardiac cycle in children typically spans 0.3 to 1 second, each pulse is resampled to ensure a uniform representation of 256 data points, equivalent to a 1-second cardiac cycle. Missing values are interpolated using a linear interpolation technique, which is selected for its computational efficiency and ability to maintain signal continuity, thereby preserving the integrity of the pulse waveform.
\item \textbf{Normalization}: The resampled data are normalized to have zero mean and unit variance, ensuring that all features in the dataset are on a comparable scale. This step prevents numerical imbalances from disproportionately influencing subsequent modeling stages.
\item \textbf{Data transformation}: Each segmented pulse is represented as a structured vector containing 256 values, corresponding to the uniformly sampled points derived in the resampling step. This standardized vector format facilitates using the PPG pulses in statistical analyses and machine learning applications, ensuring that the data are consistent and manageable for various downstream processing tasks.
\end{enumerate}

\subsection{Data Annotation}
After segmenting the PPG signals, each pulse is presented to a professional clinician at CHUSJ for annotation. The clinician classifies each pulse as either non-artifact or artifact motion. A pulse is considered artifact-free if its morphology - defined by consistent amplitude, width, and shape - matches that of surrounding pulses. Conversely, a pulse is marked as containing artifacts if it deviates significantly in these characteristics compared to its neighboring pulses.

To validate the annotations provided by the expert, an automated re-annotation algorithm, functioning as a secondary reviewer, was developed \cite{macabiau2023label}. This algorithm utilizes statistical features such as skewness, kurtosis, and standard deviation to differentiate artifact-free pulses from those containing motion artifacts. For consistent cardiac cycles, these values remain stable, while deviations indicate the presence of artifacts. Let \( X \) denote the pulse sample values, with \(\mu\) as the mean and \(\sigma\) as the standard deviation. The skewness, kurtosis, and standard deviation are defined as follows:
\begin{align}
    \text{Kurt}[X] &= \frac{E\left[\left(X-\mu\right)^4\right]}{\sigma^4} \\ 
    \text{Skew}[X] &= \frac{E\left[\left(X-\mu\right)^3\right]}{\sigma^3} \\ 
    \text{std}[X] &= \sqrt{E\left[\left(X-\mu\right)^2\right]}
\end{align}
To detect anomalies, a 95\% confidence interval is established, such that a pulse is classified as normal if its statistical values fall within:
\begin{align}
     th_l &= \mu-2\sigma\\
     th_u &= \mu+2\sigma
\end{align}
If these statistics lie outside the defined thresholds, the pulse is marked as containing artifacts. This approach helps ensure consistency and reduces the risk of human error in manual annotations, ultimately improving the reliability of the pulse classification thereby providing robust analysis.

Finally, with an average of 51 pulses per signal and 1,571 signals in the database, approximately 80,000 pulses were available. Previously, only 10\% of the pulses were annotated, leaving 90\% unannotated \cite{macabiau2023label}. To leverage this unannotated data, we will experiment with varying annotation proportions of 2.5\%, 5\%, 7.5\%, and 10\% - to determine the minimal subset required for effective artifact detection. This approach helps identify the optimal dataset size for robust performance.

\subsection{Self-Supervised Learning (SSL)}

SSL represents a breakthrough in machine learning, offering an innovative way for systems to understand and process data. Unlike traditional supervised learning that depends on human-provided labeled data, SSL generates training signals from the unlabeled data. It formulates a proxy objective, often by creating tasks where the model predicts part of the data from other parts. SSL avoids trivial solutions by employing strategies such as contrastive methods, like SimCLR and its InfoNCE criterion, which distinguish between positive and negative examples, or non-contrastive methods that apply regularization to prevent model collapse. The key advantages of SSL include its efficiency in reducing the need for extensive manual labeling, its ability to learn rich data representations beneficial for various downstream tasks, and its flexibility to be applied to diverse data types. This approach has been gaining traction, especially in fields like computer vision and natural language processing, due to its effective utilization of large volumes of unlabeled data \cite{shwartz2023information}.

Among SSL approaches, three notable techniques stand out, each offering unique approaches to training models on unlabeled data. Each of these techniques embodies the essence of SSL by extracting valuable information from unlabeled data, thereby broadening the scope and efficiency of machine learning models in various domains.

\subsubsection{Masking}

The masking approach for SSL, as depicted in the pseudo Algorithm \ref{algo:masked_data}, is a technique that involves selectively hiding parts of the input data and then training a model to predict these masked portions. This approach is widely used in various domains, such as natural language processing (e.g., BERT \cite{devlin2019bert}) and computer vision, to learn robust data representations without the need for labeled data. An explanation of the provided pseudo-algorithm is as follows: i) First, the algorithm requires the original data and a specified mask size. The output will be the masked data and the positions of the mask; ii) Then, for each row in the data, a random starting position for the mask is chosen. The range for this random integer is from 0 to the length of the row minus the mask size, ensuring the mask doesn't exceed the row boundaries. This step effectively masks or hides part of the data $masked\_data$.  The position of the mask applied is recorded in the $mask\_positions$ list; iii) Finally, once all rows have been processed, the algorithm returns the now partially masked data along with the list of mask positions.
The purpose of this algorithm is to create a scenario where the model is challenged to understand and predict the underlying structure of the data, given incomplete information. By doing so, the model learns to capture the essential features of the data, making it capable of handling similar prediction tasks. This form of SSL is powerful because it can leverage vast amounts of unlabeled data, learning from the data structure itself rather than from external annotations.

\begin{algorithm*}
\caption{Masking Data Algorithm}
\label{algo:masked_data}
\begin{algorithmic}[1]
\Require $data$, $mask\_size$
\Ensure $masked\_data$, $mask\_positions$
\State $masked\_data \gets \Call{Copy}{data}$
\State $mask\_positions \gets \text{empty list}$ 
\For{each $row$ in $data$}
    \State $mask\_position \gets \Call{RandomInteger}{0, \text{length}(row) - mask\_size}$
    \State $masked\_data[row,\ mask\_position :$ 
    \Statex\hspace{\algorithmicindent}\hspace{\algorithmicindent}\hspace{\algorithmicindent}$mask\_position + mask\_size] \gets 0$
    \State $\Call{Append}{mask\_positions, mask\_position}$
\EndFor
\State \Return $masked\_data$, $mask\_positions$
\end{algorithmic}
\end{algorithm*}

\begin{algorithm*}
\caption{A Self-Supervised Model for Data Reconstruction from the Masked Data (Algorithm \ref{algo:masked_data})}
\label{algo:self-masking}
\begin{algorithmic}[1]
\Procedure{BuildModel}{}
    \State $inputs \gets \Call{InputLayer}{shape=(feature\_size,)}$
    \State $x \gets \Call{PreprocessingLayers}{inputs}$
    \State $x \gets \Call{AddModelLayers}{x}$
    \State $outputs \gets \Call{OutputLayer}{x}$
    \State \Return $\Call{Model}{inputs, outputs}$
\EndProcedure

\State $model \gets \Call{BuildModel}{}$
\State \Call{Compile}{$model, optimizer, loss= `mse'$}
\State \Call{Fit}{$model, masked\_data, original\_data, training\_params$}
\end{algorithmic}
\end{algorithm*}

Then, the SSL framework for data reconstruction from the masked data utilizes the masked data generated by the previous pseudo-algorithm. This framework aims to train a model to predict the original data from its masked version, thereby learning a robust representation of the data. Through this procedure, Algorithm \ref{algo:self-masking}, the model learns to predict the missing values in the masked data by reconstructing the original data. As the training progresses, the model becomes better at understanding the underlying patterns and structures in the data, even when some information is obscured. This SSL training paradigm effectively leverages the data as the supervisory signal, bypassing the need for external labels and allowing the model to learn unsupervised. This is particularly powerful for utilizing large unlabeled datasets to train models for tasks where labeled data is scarce to obtain.

\subsubsection{Contrastive Learning}

Contrastive learning is an SSL technique that teaches a model to distinguish between similar and dissimilar data points \cite{le2020contrastive}. By doing so, the model learns rich, discriminative data representations without needing explicit labels. The essence of contrastive learning lies in its loss functions Eq. \ref{eq1}, \ref{eq2}, and \ref{eq3}, which drive the model to minimize the distance between positive pairs (similar items) and maximize the distance between negative pairs (dissimilar items). 
Here are three contrastive loss functions that are commonly used. Let's denote $L_{ij}$ represents the loss for a positive pair of examples $(i,j)$. $\exp\left(\text{sim}\left(\mathbf{z}_{i}, \mathbf{z}_{j}\right)/\tau\right)$ is the exponential function applied to the similarity score between the encoded representations $z_i$ and $z_j$. $\tau$ is the temperature parameter that scales the similarity score and controls the separation of the distribution of positive and negative examples.

\noindent \textbf{Normalized Temperature-scaled Cross Entropy} (NT-Xent)\cite{chen2020simple}: This loss function is central to many contrastive learning algorithms and has been popularized by its use in SimCLR. It normalizes feature vectors and scales the dot product between them with a temperature parameter, encouraging the model to identify positive pairs among negative samples.

\begin{align}
    L_{NT-Xent} = -\log \left(\frac{\exp\left(\text{sim}\left(\mathbf{z}_{i}, \mathbf{z}_{j}\right)/\tau\right)}{\sum^{2N}_{k=1} 1_{[k\neq i]}\exp\left(\text{sim}\left(\mathbf{z}_{i}, \mathbf{z}_{k}\right)/\tau\right)}\right) \label{eq1}
\end{align}

NT-Xent scales the similarity scores and, by lowering the temperature, encourages the model to focus more on "hard" negatives (i.e., those with high similarity to the anchor).

\begin{itemize}
    \item Uses the entire batch as potential negatives, making it very effective with large batch sizes.
    \item The indicator function $1_{[k\neq i]}$ ensures that the similarity of the anchor with itself is excluded from the denominator.
    \item The temperature parameter $\tau$ is used to control the separation between positives and negatives.
\end{itemize}

\begin{table*}[h!]
\centering
\caption{Comparison of Contrastive Loss Functions}
\label{tab:cont_loss_compare}
\renewcommand{\arraystretch}{1.2} 
\begin{tabularx}{\textwidth}{|c|X|X|X|X|}
\hline
\textbf{Loss Function} & \textbf{Focus} & \textbf{Denominator (Negatives)} & \textbf{Advantages} & \textbf{Disadvantages} \\ \hline
\textbf{NT-Xent}       & Identifying positive pairs in the presence of many negatives & All other samples in the batch, excluding itself. & Effective with large batch sizes. & Sensitive to batch size and temperature parameters. \\ \hline
\textbf{InfoNCE}       & Learning representations by maximizing mutual information & Explicit set of negative samples. & Strong connection to mutual information estimation. & Requires careful negative sampling. \\ \hline
\textbf{SwCE}          & Matching positive embedding to negatives & Embeddings are swapped; matches positive to other negatives. & Symmetric treatment of views can enhance robustness. & Complexity in the formulation and swapping mechanism. \\ \hline
\end{tabularx}
\end{table*}

\noindent \textbf{Information Noise-Contrastive Estimation} (InfoNCE) \cite{sohn2016improved, oord2018representation}: InfoNCE is a variant that focuses on distinguishing a positive pair from a set of negatives. It's designed to learn energy-based models and has been effectively used in representation learning, improving the quality of learned representations.
\footnotesize 
\begin{align}
    L_{InfoNCE} = -\log\left(\frac{\exp\left(\text{sim}\left(\mathbf{z}_{i}, \mathbf{z}_{j}\right)/\tau\right)}{\exp\left(\text{sim}\left(\mathbf{z}_{i}, \mathbf{z}_{j}\right)/\tau\right) + \sum\limits_{\substack{k=1}}^{N} \exp\left(\text{sim}\left(\mathbf{z}_{i}, \mathbf{z}_{k}^{neg}\right)/\tau\right)}\right) \label{eq2}
\end{align}
\normalsize

\noindent InfoNCE encourages high mutual information between the anchor and its positive pair, maximizing the similarity for the positive while minimizing similarity with negative samples.
\begin{itemize}
    \item Focuses explicitly on the positive pair and contrasts it against a predefined set of negatives.
    \item Each positive pair competes against $N$ negatives, typically resulting in better feature separation in high-dimensional spaces.
    \item The denominator sums over the positives and negatives, which ensures that the loss is bounded and smoothly differentiable.
\end{itemize}

\noindent \textbf{Swapped Cross Entropy (SwCE)}\cite{caron2020unsupervised}: Introduced in SwAV, this function enforces consistency between cluster assignments produced by a pair of encoded views of the same image. It uses a swapping mechanism that leads to more robust representations by leveraging the invariance in the data.

\begin{align}
    L_{SwCE} = -\log\left(\frac{\exp\left(\text{sim}\left(\mathbf{z}_{i}, \mathbf{z}_{j}\right)/\tau\right)}{\sum^{N}_{k=1} \exp\left(\text{sim}\left(\mathbf{z}_{j}, \mathbf{z}_{k}\right)/\tau\right)}\right) \label{eq3}
\end{align}

\noindent SwCE creates a symmetry between the views, encouraging the model to treat both views equally. This swapping mechanism can lead to more robust and balanced embeddings.
\begin{itemize}
    \item It flips the anchor-positive relationship, leading to a more balanced loss.
    \item The denominator sums over the similarities of the positive embedding ($z_j$) to all other negative embeddings.
\end{itemize}

These loss functions are integral to the SSL framework, enabling models to effectively leverage large amounts of unlabeled data to learn useful data representations. Each loss function has advantages and can be chosen based on the specific requirements summarized in the Table. \ref{tab:cont_loss_compare}. 

Within this framework, we introduce Smooth InfoNCE, an extension of the standard InfoNCE loss that aims to tackle the challenges of overfitting and sensitivity to negative samples, particularly in limited data scenarios. While the traditional InfoNCE loss treats all negative samples with equal weight, this can lead to potential overfitting and an overly confident separation between positive and negative pairs, especially when the set of negative samples is small or lacks diversity. This behavior risks the model learning to "memorize" negatives rather than capturing meaningful representations. The motivation behind the Smooth InfoNCE (Eq. \ref{eq4}), is as follows:
\begin{itemize}
    \item \textbf{Overfitting to Negatives in Limited Data Scenarios}: In standard InfoNCE, the contribution of negative samples is uniform. In cases of limited data, there is a risk of the model overfitting to these negatives. If the same small set of negatives is repeatedly used, the model might become overly confident in separating positive and negative pairs. This results in the model memorizing the negatives rather than learning meaningful representations.
    \item \textbf{Smoothing Factor to Mitigate Overconfidence:} The smoothing factor $\lambda$ in the denominator helps modulate the contribution of negative samples, thus preventing the model from becoming overly confident in its predictions. This is especially useful when the negative samples are few or noisy.
    \item \textbf{Generalization in Small Datasets:} By tuning $\lambda$, we can ensure that the loss does not emphasize the negatives, enhancing the generalization and robustness of the learned features.
\end{itemize}

\footnotesize
\begin{equation}
    L_{SInfoNCE} = -\log \left( \frac{\exp\left( \text{sim}( \mathbf{z}_i, \mathbf{z}_j ) / \tau \right)}{\exp\left( \text{sim}( \mathbf{z}_i, \mathbf{z}_j ) / \tau \right) + \lambda \sum\limits_{\substack{k=1, k \neq i}}^{N} \exp \left( \text{sim}( \mathbf{z}_i, \mathbf{z}_k ) / \tau \right)} \right) \label{eq4}
\end{equation}
\normalsize

Smooth InfoNCE Loss can be seen as a special case of the standard InfoNCE loss. Both losses share a commonality in the numerator, reflecting the similarity between an anchor point and its positive counterpart. The primary distinction between the Smooth InfoNCE loss and the standard InfoNCE loss lies in how negative samples are treated. In the standard InfoNCE loss, the denominator sums over all negative samples with a uniform contribution, which means that each negative example is given equal importance during training. This can be problematic in limited data scenarios or when the quality of negative samples varies, as it may lead to overemphasizing specific negatives and potentially cause overfitting.

In contrast, the Smooth InfoNCE loss introduces a smoothing factor $\lambda$ that scales the contribution of each negative sample. When $\lambda \leq 1$, the influence of negatives is effectively reduced, thereby preventing the model from placing excessive confidence in distinguishing positives from these specific negative examples. This change acts as a form of regularization, making the loss function more resilient to noisy or ambiguous negative samples. As a result, the model learns more balanced representations, mitigating overfitting and enhancing generalization, especially in scenarios where the set of negative samples is small, noisy, or less diverse.

\subsubsection{Self-Distillation with No Labels (DINO)}

For simplicity, DINO \cite{caron2021emerging} is illustrated in the case of one pair of views ($x_1$, $x_2$). The model passes two different random transformations of an input image to the student and teacher networks. Both networks have the same architecture but different parameters. The output of the teacher network is centered with a mean computed over the batch. Each network outputs a $K$ dimensional feature that is normalized with a temperature $softmax$ over the feature dimension. Their similarity is then measured with a cross-entropy loss. A stop-gradient operator is applied to the teacher to propagate gradients only through the student. The teacher parameters are updated with an exponential moving average of the student parameters.


\subsection{Backbone Neural Networks}
\label{sec:ml_classifiers}

Our research draws on the successes of deep learning algorithms in artifact detection within PPG signals, as evidenced by studies employing Multilayer Perceptron (MLP) and Fully Convolutional Neural Networks (FCNN) \cite{wang2017time, marzorati2022hybrid}. The effectiveness of incorporating time-domain features in deep-learning models, particularly in PPG signal analysis, has been highlighted in recent research \cite{maqsood2021benchmark}. Among these, the Bi-LSTM model, integrating time-domain features, has been noted for its superior performance in heart rate estimation across various datasets. Complementing these approaches, our team's investigation \cite{macabiau2023label} explored a range of machine learning techniques, including semi-supervised learning, conventional ML, and advanced neural networks like MLP and Transformer, specifically for detecting artifacts in PPG signals. Building on these insights, our current study will focus on benchmarking and establishing baselines using these neural network architectures. We will primarily concentrate on classifiers such as MLP, FCNN, Bi-LSTM, and Transformer, leveraging their distinct strengths in our analysis and model development.

\section{Experimental Results}
\label{sec:result_discussions}

Our experiments were carried out using the PICU e-Medical infrastructure and the Miircic Database at CHUSJ, with computational support from a GPU Quadro RTX 6000, equipped with 24 Gb of memory. For model implementation, we employed the scikit-learn library \cite{scikit-learn} and Keras \cite{chollet2015keras} within a Python environment. For each experiment, the dataset was split into 70\% for training and 30\% for evaluation.

Informed by previous studies on neural network architecture optimization \cite{hunter2012selection}, we paid special attention to the model size, learning rate, and batch size, which are crucial hyperparameters for effective Transformer model training, as described in \cite{popel2018training}. To enhance model performance and stability, we incorporated dropout \cite{srivastava2014dropout} with a probability of 0.25, and used the GlorotNormal kernel initializer \cite{glorot2010understanding} along with batch normalization \cite{ioffe2015batch, bjorck2018understanding}. Addressing the challenge of imbalanced classes, we utilized the ADASYN method \cite{he2008adasyn} for oversampling. These hyperparameters were meticulously selected to ensure optimal performance while mitigating the risk of overfitting.

To effectively evaluate the performance of our method, we utilized several key metrics: accuracy, precision, recall (or sensitivity), and the F1 score. These metrics are crucial for comprehensively assessing our model's performance. The formulas for these metrics are as follows:

\begin{align}
&\text {Accuracy (acc) }=\frac{\mathrm{TP}+\mathrm{TN}}{\mathrm{TP}+\mathrm{TN}+\mathrm{FP}+\mathrm{FN}} \nonumber \\ 
&\text {Precision (pre) }=\frac{\mathrm{TP}}{\mathrm{TP}+\mathrm{FP}} \nonumber \\
&\text {Recall/Sensitivity (rec)}=\frac{\mathrm{TP}}{\mathrm{TP}+\mathrm{FN}} \nonumber \\ 
&\text {F1-Score (f1)} =\frac{2^{\star} \text {Precision}^{\star} \text {Recall}}{\text {Precision }+\text {Recall}} \nonumber
\end{align}

In these metrics, TN (True Negative) and TP (True Positive) refer to the number of correctly classified negative and positive cases, respectively. Specifically, TN denotes cases that are correctly identified as negative, and TP indicates cases accurately recognized as positive. On the other hand, FP (False Positive) and FN (False Negative) represent the instances of incorrect predictions. FP occurs when a negative case is incorrectly predicted as positive, while FN happens when a positive case is wrongly classified as negative. These four elements are pivotal in assessing the model's ability to distinguish between positive and negative cases accurately.

\begin{table*}[]
\centering
\caption{Parameters setting of classifiers}
\label{tab:my_hyperparameters}
\begin{tabular}{|l|l|l|l|l|}
\hline
Hyperparameters                   & Transformer & LSTM     & FCNN     & MLP        \\ \hline
Hidden layers                     & 4           & 2        & 3        & 3            \\ \hline
Number of   neurons               & 128         & 500      & 64       & 500        \\ \hline
Number of   multi-heads attention & 4           & N/A      & N/A      & N/A        \\ \hline
Batch size                        & 96          & 96       & 96       & 96         \\ \hline
Dropout                           & 0.25        & 0.3      & 0.25     & 0.3       \\ \hline
Learning rate                     & 6e-04       & 1e-04     & 1e-04     & 1e-04   \\ \hline
Optimizer                         & Adam        & Adam     & Adam     & Adam       \\ \hline
\end{tabular}
\end{table*}

Table \ref{tab:my_hyperparameters} summarizes the hyperparameters used for different ML classifiers. The Transformer has 4 hidden layers with 128 neurons each, 4 multi-head attention mechanisms, a batch size of 96, a dropout rate of 0.25, a learning rate of 6e-04, and uses the Adam optimizer. The LSTM has two hidden layers with 500 neurons each, no multi-head attention, and a learning rate 1e-04. The FCNN and MLP classifiers share settings similar to those of the LSTM, including batch size, dropout rate, learning rate, and optimizer. Consequently, one of the importance of hyperparameters tuning of InfoNCE should be noted as follows:
\begin{itemize}
    \item \textbf{Temperature Parameter} (\(\tau\)): The temperature controls the sharpness of the softmax distribution. A low value (e.g., 0.1) makes the distribution sharper and emphasizes the most similar pairs more, while a higher value smooths the distribution.
    \item \textbf{Smoothing Factor} (\(\lambda\)): The \(\lambda\) value should be selected based on the dataset size and nature of the negative samples. Start with \(\lambda = 0.5\) and increase/decrease it to observe the effect. A lower \(\lambda\) will reduce the impact of negatives, which is helpful if the negatives are noisy or very similar to positives.
\end{itemize}

\begin{figure*}[!htp]
	\centering
	\includegraphics[scale=0.35]{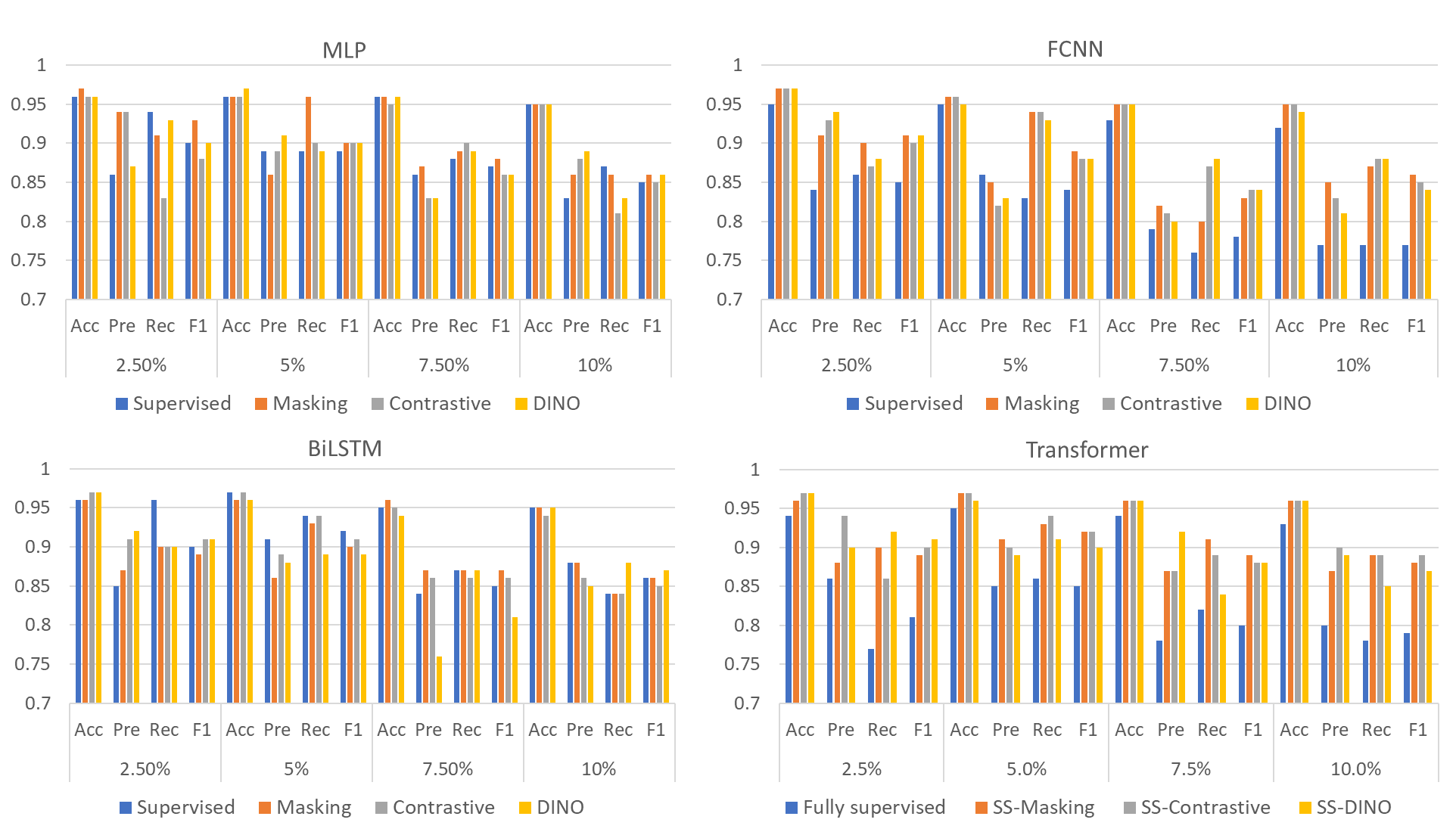}
	\caption{Classifier's performance with different learning frameworks including fully-supervised, self-supervised learning with masking, contrastive, and DINO.}
	\label{fig:performance_comparison}
\end{figure*}

\begin{table*}[]
\caption{A comparison of self-supervised learning with different backbone}
\label{tab:bacbon_comparision}
\begin{tabular}{|cl||cccc||cccc||cccc||cccc|}
\hline
\multicolumn{2}{|c|}{} &
  \multicolumn{4}{c||}{2.5\%} &
  \multicolumn{4}{c||}{5\%} &
  \multicolumn{4}{c||}{7.5\%} &
  \multicolumn{4}{c|}{10\%} \\ \cline{3-18} 
\multicolumn{2}{|c|}{\multirow{-2}{*}{Models}} &
  \multicolumn{1}{l|}{Acc} &
  \multicolumn{1}{l|}{Pre} &
  \multicolumn{1}{l|}{Rec} &
  F1 &
  \multicolumn{1}{l|}{Acc} &
  \multicolumn{1}{l|}{Pre} &
  \multicolumn{1}{l|}{Rec} &
  F1 &
  \multicolumn{1}{l|}{Acc} &
  \multicolumn{1}{l|}{Pre} &
  \multicolumn{1}{l|}{Rec} &
  F1 &
  \multicolumn{1}{l|}{Acc} &
  \multicolumn{1}{l|}{Pre} &
  \multicolumn{1}{l|}{Rec} &
  F1 \\ \hline
\multicolumn{1}{|c|}{} &
  {MLP} &
  \multicolumn{1}{l|}{{0.96}} &
  \multicolumn{1}{l|}{{0.86}} &
  \multicolumn{1}{l|}{{0.94}} &
  {{0.90}} &
  \multicolumn{1}{l|}{{0.96}} &
  \multicolumn{1}{l|}{{0.89}} &
  \multicolumn{1}{l|}{{0.89}} &
  {{0.89}} &
  \multicolumn{1}{l|}{{0.96}} &
  \multicolumn{1}{l|}{{0.86}} &
  \multicolumn{1}{l|}{{0.88}} &
  {{0.87}} &
  \multicolumn{1}{l|}{{0.95}} &
  \multicolumn{1}{l|}{{0.83}} &
  \multicolumn{1}{l|}{{0.87}} &
  {{0.85}} \\ \cline{2-18} 
\multicolumn{1}{|c|}{} &
  FCNN &
  \multicolumn{1}{l|}{{0.95}} &
  \multicolumn{1}{l|}{{0.84}} &
  \multicolumn{1}{l|}{{0.86}} &
  {{0.85}} &
  \multicolumn{1}{l|}{{0.95}} &
  \multicolumn{1}{l|}{{0.86}} &
  \multicolumn{1}{l|}{{0.83}} &
  {{0.84}} &
  \multicolumn{1}{l|}{{0.93}} &
  \multicolumn{1}{l|}{{0.79}} &
  \multicolumn{1}{l|}{{0.76}} &
  {{0.78}} &
  \multicolumn{1}{l|}{0.92} &
  \multicolumn{1}{l|}{0.77} &
  \multicolumn{1}{l|}{0.77} &
  {{0.77}} \\ \cline{2-18} 
\multicolumn{1}{|c|}{} &
  {BiLSTM} &
  \multicolumn{1}{c|}{{0.96}} &
  \multicolumn{1}{c|}{{0.85}}&
  \multicolumn{1}{c|}{{0.96}} &
  {{0.90}} &
  \multicolumn{1}{c|}{{0.97}} &
  \multicolumn{1}{c|}{{0.91}} &
  \multicolumn{1}{c|}{{0.94}} &
  {{0.92}} &
  \multicolumn{1}{c|}{{0.95}} &
  \multicolumn{1}{c|}{{0.84}} &
  \multicolumn{1}{c|}{{0.87}} &
  {{0.85}} &
  \multicolumn{1}{c|}{{0.95}} &
  \multicolumn{1}{c|}{{0.88}} &
  \multicolumn{1}{c|}{{0.84}} &
  {{0.86}} \\ \cline{2-18} 
\multicolumn{1}{|c|}{\multirow{-4}{*}{\rot{Supervised}}} &
  {Transformer} &
  \multicolumn{1}{l|}{{\textbf{0.94}}} &
  \multicolumn{1}{l|}{{\textbf{0.86}}} &
  \multicolumn{1}{l|}{{\textbf{0.77}}} &
  {\textbf{0.81}} &
  \multicolumn{1}{l|}{{\textbf{0.95}}} &
  \multicolumn{1}{l|}{{\textbf{0.85}}} &
  \multicolumn{1}{l|}{{\textbf{0.86}}} &
  {\textbf{0.85}} &
  \multicolumn{1}{l|}{{0.94}} &
  \multicolumn{1}{l|}{{ 0.78}} &
  \multicolumn{1}{l|}{{\textbf{0.82}}} &
  {\textbf{0.80}} &
  \multicolumn{1}{l|}{{\textbf{0.93}}} &
  \multicolumn{1}{l|}{{0.80}} &
  \multicolumn{1}{l|}{{\textbf{0.78}}} &
  {0.79} \\ \hline \hline
\multicolumn{1}{|c|}{} &
  MLP &
  \multicolumn{1}{l|}{{0.97}} &
  \multicolumn{1}{l|}{0.94} &
  \multicolumn{1}{l|}{0.91} &
  {0.93} &
  \multicolumn{1}{l|}{0.96} &
  \multicolumn{1}{l|}{0.86} &
  \multicolumn{1}{l|}{0.96} &
  0.90 &
  \multicolumn{1}{l|}{{0.96}} &
  \multicolumn{1}{l|}{{0.87}} &
  \multicolumn{1}{l|}{0.89} &
  0.88 &
  \multicolumn{1}{l|}{0.95} &
  \multicolumn{1}{l|}{0.86} &
  \multicolumn{1}{l|}{0.86} &
  0.86 \\ \cline{2-18} 
\multicolumn{1}{|c|}{} &
  FCNN &
  \multicolumn{1}{l|}{{0.97}} &
  \multicolumn{1}{l|}{0.91} &
  \multicolumn{1}{l|}{0.90} &
  {0.91} &
  \multicolumn{1}{l|}{0.96} &
  \multicolumn{1}{l|}{0.85} &
  \multicolumn{1}{l|}{0.94} &
  0.89 &
  \multicolumn{1}{l|}{0.95} &
  \multicolumn{1}{l|}{0.82} &
  \multicolumn{1}{l|}{0.80} &
  0.83 &
  \multicolumn{1}{l|}{0.95} &
  \multicolumn{1}{l|}{0.85} &
  \multicolumn{1}{l|}{0.87} &
  0.86 \\ \cline{2-18} 
\multicolumn{1}{|c|}{} &
  BiLSTM &
  \multicolumn{1}{c|}{0.96} &
  \multicolumn{1}{c|}{0.87} &
  \multicolumn{1}{c|}{0.90} &
  {0.89} &
  \multicolumn{1}{c|}{0.96} &
  \multicolumn{1}{c|}{0.86} &
  \multicolumn{1}{c|}{0.93} &
  {0.90} &
  \multicolumn{1}{c|}{0.96} &
  \multicolumn{1}{c|}{0.87} &
  \multicolumn{1}{c|}{0.87} &
  {0.87} &
  \multicolumn{1}{c|}{0.95} &
  \multicolumn{1}{c|}{0.88} &
  \multicolumn{1}{c|}{0.84} &
  {0.86} \\ \cline{2-18} 
\multicolumn{1}{|c|}{\multirow{-4}{*}{\rot{Masking}}} &
  Transformer &
  \multicolumn{1}{l|}{0.96} &
  \multicolumn{1}{l|}{0.88} &
  \multicolumn{1}{l|}{0.90} &
  0.89 &
  \multicolumn{1}{l|}{0.97} &
  \multicolumn{1}{l|}{0.91} &
  \multicolumn{1}{l|}{0.93} &
  {0.92} &
  \multicolumn{1}{l|}{0.96} &
  \multicolumn{1}{l|}{0.87} &
  \multicolumn{1}{l|}{0.91} &
  {0.89} &
  \multicolumn{1}{l|}{0.96} &
  \multicolumn{1}{l|}{0.87} &
  \multicolumn{1}{l|}{0.89} &
  0.88 \\ \hline\hline
\multicolumn{1}{|c|}{} &
  MLP &
  \multicolumn{1}{l|}{0.96} &
  \multicolumn{1}{l|}{{0.94}} &
  \multicolumn{1}{l|}{0.83} &
  {0.88} &
  \multicolumn{1}{l|}{0.96} &
  \multicolumn{1}{l|}{0.89} &
  \multicolumn{1}{l|}{0.9} &
  0.90 &
  \multicolumn{1}{l|}{0.95} &
  \multicolumn{1}{l|}{0.83} &
  \multicolumn{1}{l|}{0.9} &
  0.86 &
  \multicolumn{1}{l|}{0.95} &
  \multicolumn{1}{l|}{0.88} &
  \multicolumn{1}{l|}{0.81} &
  0.85 \\ \cline{2-18} 
\multicolumn{1}{|c|}{} &
  FCNN &
  \multicolumn{1}{l|}{0.97} &
  \multicolumn{1}{l|}{0.93} &
  \multicolumn{1}{l|}{0.87} &
  0.90 &
  \multicolumn{1}{l|}{0.96} &
  \multicolumn{1}{l|}{0.82} &
  \multicolumn{1}{l|}{0.94} &
  0.88 &
  \multicolumn{1}{l|}{0.95} &
  \multicolumn{1}{l|}{0.81} &
  \multicolumn{1}{l|}{0.87} &
  0.84 &
  \multicolumn{1}{l|}{0.95} &
  \multicolumn{1}{l|}{0.83} &
  \multicolumn{1}{l|}{0.88} &
  0.85 \\ \cline{2-18} 
\multicolumn{1}{|c|}{} &
  BiLSTM &
  \multicolumn{1}{l|}{{{0.97}}} &
  \multicolumn{1}{l|}{{0.91}} &
  \multicolumn{1}{l|}{{0.90}} &
  {{0.91}} &
  \multicolumn{1}{c|}{0.97} &
  \multicolumn{1}{c|}{0.89} &
  \multicolumn{1}{c|}{0.94} &
  {0.91} &
  \multicolumn{1}{c|}{{0.95}} &
  \multicolumn{1}{c|}{{0.86}} &
  \multicolumn{1}{c|}{{0.86}} &
  {{0.86}} &
  \multicolumn{1}{c|}{{0.94}} &
  \multicolumn{1}{c|}{{0.86}} &
  \multicolumn{1}{c|}{{0.84}} &
  {{0.85}} \\ \cline{2-18} 
\multicolumn{1}{|c|}{\multirow{-4}{*}{\rot{Contrastive}}} &
  Transformer &
  \multicolumn{1}{l|}{{0.97}} &
  \multicolumn{1}{l|}{{0.94}} &
  \multicolumn{1}{l|}{0.86} &
  {0.90} &
  \multicolumn{1}{l|}{{0.97}} &
  \multicolumn{1}{l|}{{0.90}} &
  \multicolumn{1}{l|}{{0.94}} &
  {0.92} &
  \multicolumn{1}{l|}{{0.96}} &
  \multicolumn{1}{l|}{0.87} &
  \multicolumn{1}{l|}{{0.89}} &
  {0.88} &
  \multicolumn{1}{l|}{{0.96}} &
  \multicolumn{1}{l|}{0.90} &
  \multicolumn{1}{l|}{{0.89}} &
  {0.89} \\ \hline\hline
\multicolumn{1}{|c|}{} &
  MLP &
  \multicolumn{1}{l|}{0.96} &
  \multicolumn{1}{l|}{0.87} &
  \multicolumn{1}{l|}{{0.93}} &
  0.90 &
  \multicolumn{1}{l|}{{0.97}} &
  \multicolumn{1}{l|}{{0.91}} &
  \multicolumn{1}{l|}{0.89} &
  {0.90} &
  \multicolumn{1}{l|}{{0.96}} &
  \multicolumn{1}{l|}{0.83} &
  \multicolumn{1}{l|}{{0.89}} &
  0.86 &
  \multicolumn{1}{l|}{0.95} &
  \multicolumn{1}{l|}{0.89} &
  \multicolumn{1}{l|}{0.83} &
  0.86 \\ \cline{2-18} 
\multicolumn{1}{|c|}{} &
  FCNN &
  \multicolumn{1}{l|}{{0.97}} &
  \multicolumn{1}{l|}{{0.94}} &
  \multicolumn{1}{l|}{0.88} &
  {0.91} &
  \multicolumn{1}{l|}{0.95} &
  \multicolumn{1}{l|}{0.83} &
  \multicolumn{1}{l|}{0.93} &
  0.88 &
  \multicolumn{1}{l|}{0.95} &
  \multicolumn{1}{l|}{0.80} &
  \multicolumn{1}{l|}{0.88} &
  0.84 &
  \multicolumn{1}{l|}{0.94} &
  \multicolumn{1}{l|}{0.81} &
  \multicolumn{1}{l|}{{0.88}} &
  0.84 \\ \cline{2-18} 
\multicolumn{1}{|c|}{} &
  BiLSTM &
  \multicolumn{1}{l|}{{0.97}} &
  \multicolumn{1}{l|}{0.92} &
  \multicolumn{1}{l|}{0.90} &
  {0.91} &
  \multicolumn{1}{l|}{0.96} &
  \multicolumn{1}{l|}{0.88} &
  \multicolumn{1}{l|}{0.89} &
  0.89 &
  \multicolumn{1}{l|}{{0.94}} &
  \multicolumn{1}{l|}{0.76} &
  \multicolumn{1}{l|}{0.87} &
  0.81 &
  \multicolumn{1}{l|}{0.95} &
  \multicolumn{1}{l|}{0.85} &
  \multicolumn{1}{l|}{{0.88}} &
  {0.87} \\ \cline{2-18} 
\multicolumn{1}{|c|}{\multirow{-4}{*}{\rot{DINO}}} &
  Transformer &
  \multicolumn{1}{l|}{{0.97}} &
  \multicolumn{1}{l|}{0.90} &
  \multicolumn{1}{l|}{0.92} &
  {0.91} &
  \multicolumn{1}{l|}{0.96} &
  \multicolumn{1}{l|}{0.89} &
  \multicolumn{1}{l|}{{0.91}} &
  {0.90} &
  \multicolumn{1}{l|}{{0.96}} &
  \multicolumn{1}{l|}{0.92} &
  \multicolumn{1}{l|}{0.84} &
  {0.88} &
  \multicolumn{1}{l|}{{0.96}} &
  \multicolumn{1}{l|}{0.89} &
  \multicolumn{1}{l|}{0.85} &
  {0.87} \\ \hline
\end{tabular}
\end{table*}

The table \ref{tab:bacbon_comparision} comprehensively evaluates various machine learning models across different learning paradigms and data proportions. It compares the performance of MLP, FCNN, BiLSTM, and Transformer models, each trained under fully supervised, masking, contrastive, and DINO self-supervised learning paradigms. The performance metrics include Accuracy (Acc), Precision (Pre), Recall (Rec), and F1 Score (F1), evaluated at four different ratios of annotated data: 2.5\%, 5\%, 7.5\%, and 10\%. Table \ref{tab:bacbon_comparision} shows that the Transformer model, particularly when trained with SSL approaches like masking, contrastive, and DINO, consistently achieves high scores across all metrics, indicating its robustness and adaptability in learning from annotated and unannotated data. It significantly improves the Transformer's performance compared to supervised learning cases. For example, under the DINO self-supervised paradigm, the Transformer model shows remarkable performance with an accuracy and F1 score of 0.97 and 0.91, respectively, at the 2.5\% data ratio and maintains similar high performance across other data ratios. In contrast, while MLP and BiLSTM models perform best under fully supervised learning, their performance does not significantly improve with SSL techniques. The FCNN model, however, shows some improvement with SSL compared to fully supervised learning, but not to the extent seen with the Transformer model. Overall, these results underline the effectiveness of SSL, particularly with the Transformer model, in handling various proportions of annotated and unannotated data, outperforming traditional supervised methods and other neural network architectures like MLP, BiLSTM, and FCNN in terms of all evaluation metrics.

Moreover, from the visual data in Fig. \ref{fig:performance_comparison}, it is clear that the Transformer model consistently outperforms other models in SSL scenarios across all metrics and data proportions. Particularly under the contrastive paradigm, the Transformer model shows robust performance, often reaching peak scores. In contrast, while MLP and BiLSTM models exhibit strong performance in supervised learning, they do not exhibit the same improvement with SSL as the Transformer. FCNN shows a moderate improvement with SSL, suggesting that it benefits from these paradigms, but not as significantly as the Transformer. The trend across all models indicates that SSL, especially with the contrastive approach, substantially enhances model performance, with the Transformer model achieving notable improvements in learning from unannotated data and fine-tuning on annotated data. The bar charts serve as a clear visual testament to the advantages of employing SSL techniques in enhancing model robustness, with the Transformer model standing out as a particular architecture.

\begin{figure*}[!htp]
	\centering
	\includegraphics[scale=0.325]{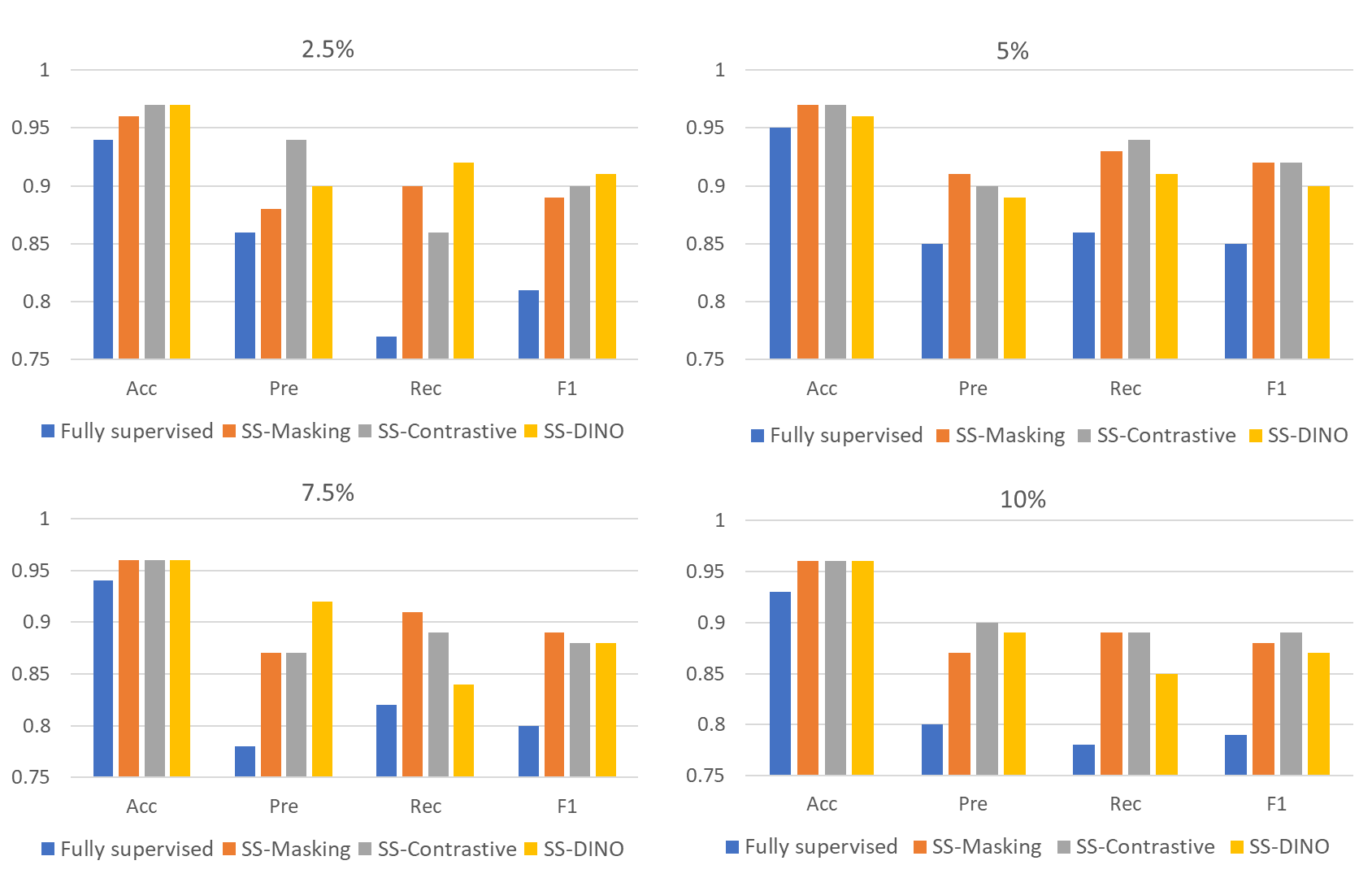}
	\caption{Transformer's performance with different learning frameworks including fully-supervised, self-supervised learning with masking, contrastive, and DINO.}
	\label{fig:performance_transformer}
\end{figure*}

\begin{table*}[]
\caption{Transformer with self-supervised learning regarding different frameworks}
\label{tab:ssl_comparision}
\begin{tabular}{|l||llll||llll||llll||llll|}
\hline
\multicolumn{1}{|c|}{} &
  \multicolumn{4}{c||}{2.5\%} &
  \multicolumn{4}{c||}{5\%} &
  \multicolumn{4}{c||}{7.5\%} &
  \multicolumn{4}{c|}{10\%} \\ \cline{2-17} 
\multicolumn{1}{|c|}{\multirow{-2}{*}{Models}} &
  \multicolumn{1}{l|}{Acc} &
  \multicolumn{1}{l|}{Pre} &
  \multicolumn{1}{l|}{Rec} &
  F1 &
  \multicolumn{1}{l|}{Acc} &
  \multicolumn{1}{l|}{Pre} &
  \multicolumn{1}{l|}{Rec} &
  F1 &
  \multicolumn{1}{l|}{Acc} &
  \multicolumn{1}{l|}{Pre} &
  \multicolumn{1}{l|}{Rec} &
  F1 &
  \multicolumn{1}{l|}{Acc} &
  \multicolumn{1}{l|}{Pre} &
  \multicolumn{1}{l|}{Rec} &
  F1 \\ \hline
Fully   supervised &
  \multicolumn{1}{l|}{{0.94}} &
  \multicolumn{1}{l|}{{0.86}} &
  \multicolumn{1}{l|}{{0.77}} &
  {0.81} &
  \multicolumn{1}{l|}{{0.95}} &
  \multicolumn{1}{l|}{{0.85}} &
  \multicolumn{1}{l|}{{0.86}} &
  {0.85} &
  \multicolumn{1}{l|}{{0.94}} &
  \multicolumn{1}{l|}{{0.78}} &
  \multicolumn{1}{l|}{{0.82}} &
  {0.80} &
  \multicolumn{1}{l|}{{0.93}} &
  \multicolumn{1}{l|}{{0.80}} &
  \multicolumn{1}{l|}{{ 0.78}} &
  {0.79} \\ \hline
SS-Masking &
  \multicolumn{1}{l|}{0.96} &
  \multicolumn{1}{l|}{0.88} &
  \multicolumn{1}{l|}{0.90} &
  0.89 &
  \multicolumn{1}{l|}{\textbf{0.97}} &
  \multicolumn{1}{l|}{\textbf{0.91}} &
  \multicolumn{1}{l|}{0.93} &
  \textbf{0.92} &
  \multicolumn{1}{l|}{\textbf{0.96}} &
  \multicolumn{1}{l|}{0.87} &
  \multicolumn{1}{l|}{\textbf{0.91}} &
  \textbf{0.89} &
  \multicolumn{1}{l|}{\textbf{0.96}} &
  \multicolumn{1}{l|}{0.87} &
  \multicolumn{1}{l|}{0.89} &
  0.88 \\ \hline
SS-Contrastive &
  \multicolumn{1}{l|}{\textbf{0.97}} &
  \multicolumn{1}{l|}{\textbf{0.94}} &
  \multicolumn{1}{l|}{0.86} &
  0.90 &
  \multicolumn{1}{l|}{\textbf{0.97}} &
  \multicolumn{1}{l|}{0.90} &
  \multicolumn{1}{l|}{\textbf{0.94}} &
  \textbf{0.92} &
  \multicolumn{1}{l|}{\textbf{0.96}} &
  \multicolumn{1}{l|}{0.87} &
  \multicolumn{1}{l|}{0.89} &
  0.88 &
  \multicolumn{1}{l|}{\textbf{0.96}} &
  \multicolumn{1}{l|}{\textbf{0.90}} &
  \multicolumn{1}{l|}{\textbf{0.89}} &
  \textbf{0.89} \\ \hline
SS-DINO &
  \multicolumn{1}{l|}{\textbf{0.97}} &
  \multicolumn{1}{l|}{0.90} &
  \multicolumn{1}{l|}{\textbf{0.92}} &
  \textbf{0.91} &
  \multicolumn{1}{l|}{0.96} &
  \multicolumn{1}{l|}{0.89} &
  \multicolumn{1}{l|}{0.91} &
  0.90 &
  \multicolumn{1}{l|}{\textbf{0.96}} &
  \multicolumn{1}{l|}{\textbf{0.92}} &
  \multicolumn{1}{l|}{0.84} &
  0.88 &
  \multicolumn{1}{l|}{\textbf{0.96}} &
  \multicolumn{1}{l|}{0.89} &
  \multicolumn{1}{l|}{0.85} &
  0.87 \\ \hline
\end{tabular}
\end{table*}

Table \ref{tab:ssl_comparision} showcases a comparative analysis of the Transformer model's efficacy when trained under different SSL frameworks—masking, contrastive, and DINO—relative to traditional fully supervised training. At the outset, with only 2.5\% annotated data, the SSL models generally surpass the fully supervised model on all counts. The contrastive and DINO approaches, in particular, demonstrate marked improvements, with the contrastive framework achieving top scores for accuracy and F1. As we escalate the annotated data to 5\% and 7.5\%, the contrastive learning framework notably maintains its high scores, aligning with the DINO approach in terms of accuracy and F1, and consistently outshining the fully supervised model. 
Upon reaching 10\% annotated data, this pattern persists, with SSL frameworks, especially the contrastive and DINO, proving more efficient than fully supervised learning. They register high F1, indicative of their capacity to effectively harness annotated and unannotated data.

Fig. \ref{fig:performance_transformer} complements these findings with a visual representation, further illustrating the superiority of SSL, particularly in the contrastive learning and DINO variants, across varying levels of annotated data. Even with a mere 2.5\% of data annotated, these SSL methods demonstrate significant enhancements in model performance, a trend that is sustained as the volume of annotated data grows. The contrastive approach, in particular, excels across the board, while DINO shows a strong balance between precision and recall, a critical factor in robust model training. The cumulative evidence from the table and figure underscores the robustness and efficiency of SSL frameworks, with the contrastive and DINO methods standing out. These SSL techniques not only improve the Transformer model's performance in data-scarce situations but also exhibit remarkable adaptability and reliability across varied data availability scenarios, emphasizing the strengths of SSL in enhancing model performance.

\begin{table*}[]
\caption{A comparison of contrastive loss from contrastive learning with Transformer}
\label{tab:contrastiveloss_comparision}
\begin{tabular}{|l||llll||llll||llll||llll|}
\hline
\multicolumn{1}{|c|}{} &
  \multicolumn{4}{c||}{2.5\%} &
  \multicolumn{4}{c||}{5\%} &
  \multicolumn{4}{c||}{7.5\%} &
  \multicolumn{4}{c|}{10\%} \\ \cline{2-17} 
\multicolumn{1}{|c|}{\multirow{-2}{*}{Contrastive Loss}} &
  \multicolumn{1}{l|}{Acc} &
  \multicolumn{1}{l|}{Pre} &
  \multicolumn{1}{l|}{Rec} &
  F1 &
  \multicolumn{1}{l|}{Acc} &
  \multicolumn{1}{l|}{Pre} &
  \multicolumn{1}{l|}{Rec} &
  F1 &
  \multicolumn{1}{l|}{Acc} &
  \multicolumn{1}{l|}{Pre} &
  \multicolumn{1}{l|}{Rec} &
  F1 &
  \multicolumn{1}{l|}{Acc} &
  \multicolumn{1}{l|}{Pre} &
  \multicolumn{1}{l|}{Rec} &
  F1 \\ \hline
NT-Xent &
  \multicolumn{1}{l|}{\textbf{0.97}} &
  \multicolumn{1}{l|}{{\textbf{0.94}}} &
  \multicolumn{1}{l|}{{0.86}} &
  {0.90} &
  \multicolumn{1}{l|}{{0.97}} &
  \multicolumn{1}{l|}{{0.90}} &
  \multicolumn{1}{l|}{{\textbf{0.94}}} &
  {0.92} &
  \multicolumn{1}{l|}{{0.96}} &
  \multicolumn{1}{l|}{{0.87}} &
  \multicolumn{1}{l|}{{0.89}} &
  {0.88} &
  \multicolumn{1}{l|}{{\textbf{0.96}}} &
  \multicolumn{1}{l|}{{0.90}} &
  \multicolumn{1}{l|}{{\textbf{0.89}}} &
  {\textbf{0.89}} \\ \hline
SwCE &
  \multicolumn{1}{l|}{\t{0.97}} &
  \multicolumn{1}{l|}{0.92} &
  \multicolumn{1}{l|}{0.89} &
  0.90 &
  \multicolumn{1}{l|}{{0.97}} &
  \multicolumn{1}{l|}{{0.90}} &
  \multicolumn{1}{l|}{\textbf{0.94}} &
  {0.92} &
  \multicolumn{1}{l|}{\textbf{0.97}} &
  \multicolumn{1}{l|}{{0.89}} &
  \multicolumn{1}{l|}{0.89} &
  {0.89} &
  \multicolumn{1}{l|}{0.95} &
  \multicolumn{1}{l|}{0.90} &
  \multicolumn{1}{l|}{0.83} &
  0.86 \\ \hline
InfoNCE &
  \multicolumn{1}{c|}{{0.96}} &
  \multicolumn{1}{c|}{{0.90}} &
  \multicolumn{1}{c|}{\textbf{0.90}} &
  {{0.90}} &
  \multicolumn{1}{c|}{{0.97}} &
  \multicolumn{1}{c|}{{0.91}} &
  \multicolumn{1}{c|}{{0.93}} &
  {{0.92}} &
  \multicolumn{1}{c|}{{0.96}} &
  \multicolumn{1}{c|}{0.87} &
  \multicolumn{1}{c|}{{0.89}} &
  {{0.88}} &
  \multicolumn{1}{c|}{\textbf{0.96}} &
  \multicolumn{1}{c|}{0.88} &
  \multicolumn{1}{c|}{{0.87}} &
 {{0.87}} \\ \hline
Smooth InfoNCE &
  \multicolumn{1}{l|}{\textbf{0.97}} &
  \multicolumn{1}{l|}{\textbf{0.94}} &
  \multicolumn{1}{l|}{\textbf{0.90}} &
  \textbf{0.93} &
  \multicolumn{1}{l|}{{0.97}} &
  \multicolumn{1}{l|}{\textbf{0.93}} &
  \multicolumn{1}{l|}{{0.92}} &
  \textbf{0.93} &
  \multicolumn{1}{l|}{\textbf{0.97}} &
  \multicolumn{1}{l|}{\textbf{0.90}} &
  \multicolumn{1}{l|}{\textbf{0.90}} &
  {\textbf{0.90}} &
  \multicolumn{1}{l|}{\textbf{0.96}} &
  \multicolumn{1}{l|}{\textbf{0.92}} &
  \multicolumn{1}{l|}{\textbf{0.89}} &
  {\textbf{0.90}} \\ \hline
\end{tabular}
\end{table*}

\begin{figure*}[!htp]
	\centering
	\includegraphics[scale=0.425]{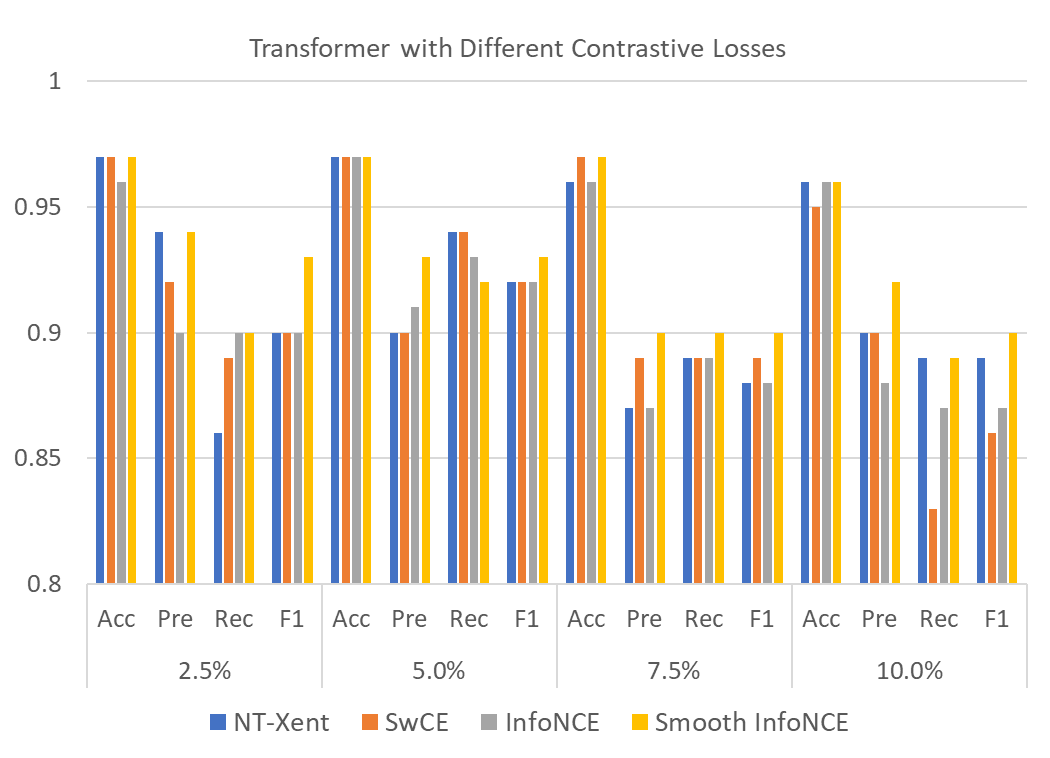}
	\caption{Contrastive self-supervised learning with different contrastive losses for Transformer.}
	\label{fig:ssl_transformers}
\end{figure*}

Table \ref{tab:contrastiveloss_comparision} evaluates the Transformer model using contrastive loss functions, including NT-Xent, SwCE, InfoNCE, and Smooth InfoNCE, revealing that with just 2.5\% annotated data, NT-Xent and Smooth InfoNCE excel in accuracy and F1 score. As the annotated portion increases to 5\% and 7.5\%, these losses, along with SwCE, deliver top F1 scores, demonstrating a balanced precision-recall trade-off. At 10\% annotation, they continue to showcase high accuracy and precision, with Smooth InfoNCE maintaining a notable F1 score. Synthesizing these insights, it is evident that contrastive loss functions significantly enhance the Transformer model's performance, especially when annotated data is scarce. NT-Xent and Smooth InfoNCE, in particular, demonstrate consistent strength across all metrics, underlining their effectiveness within the Transformer's contrastive learning framework.

Complementing Table \ref{tab:contrastiveloss_comparision}, Figure \ref{fig:ssl_transformers} graphically shows the performance of the Transformer model under various contrastive losses. At the 2.5\% annotation level, NT-Xent and Smooth InfoNCE manifest as superior performers. As the proportion of annotated data increases, these two losses retain their prominence, suggesting their efficacy in contexts with limited annotations. By the time annotated data reaches 10\%, performances across the different loss functions begin to blend, with Smooth InfoNCE maintaining its strong precision, which could indicate a particular proficiency in generating confident predictions. Those results highlight the efficacy of contrastive loss functions in training Transformers, particularly in data-constrained environments. The consistently high performance of Smooth InfoNCE and NT-Xent suggests these methods are especially potent, showcasing the transformative potential of contrastive learning to boost Transformer models when resources are limited.

\begin{figure*}[!htp]
	\centering
	\includegraphics[scale=0.6]{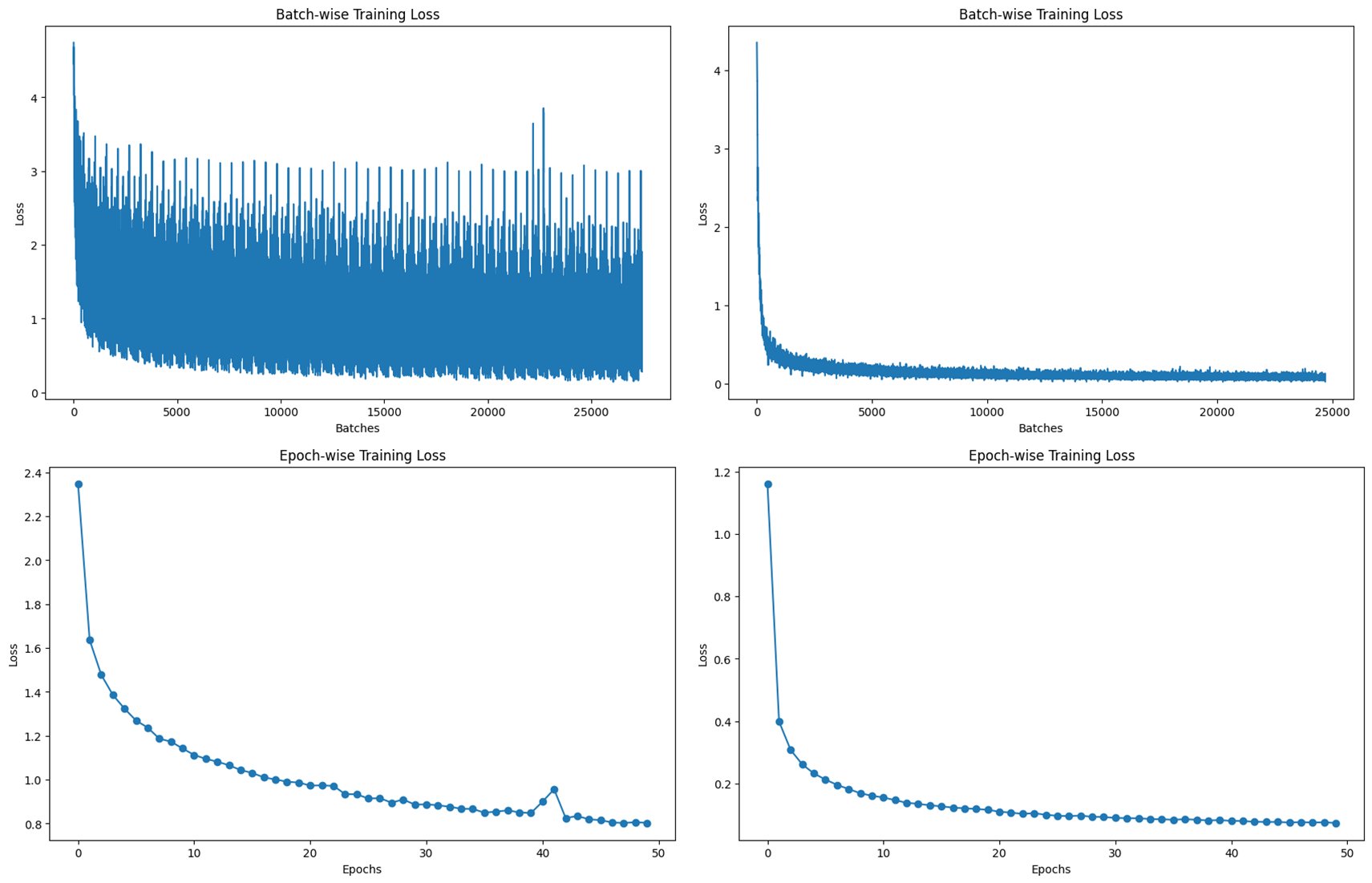}
	\caption{Training loss with InfoNCE (left) and SmoothInfoNCE (right) contrastive loss regarding to batches and epochs.}
	\label{fig:infonce_comparison}
\end{figure*}

Additionally, Figure \ref{fig:infonce_comparison} presents a side-by-side comparison of the training loss for models utilizing InfoNCE and Smooth InfoNCE contrastive losses, with the left side depicting InfoNCE and the right side Smooth InfoNCE (with the fine-tunned $\lambda = 0.75$). In the batch-wise training loss comparison, the InfoNCE loss exhibits high initial values and substantial fluctuation throughout training, suggesting some instability. Conversely, the Smooth InfoNCE demonstrates a quick reduction in loss, maintaining a lower and more stable trajectory, indicative of a steadier learning process. Epoch-wise, the InfoNCE loss graph shows an initial steep decline that gradually levels off, albeit with some irregularities that hint at ongoing learning adjustments. The Smooth InfoNCE's epoch-wise loss, however, decreases consistently and without interruption, reflecting a more uniform and presumably more effective optimization over time. This overall comparison indicates that Smooth InfoNCE provides a more controlled and consistent reduction in training loss, both within batches and across epochs, potentially leading to better model generalization and performance stability. 

\begin{figure*}[!htp]
	\centering
	\includegraphics[scale=0.475]{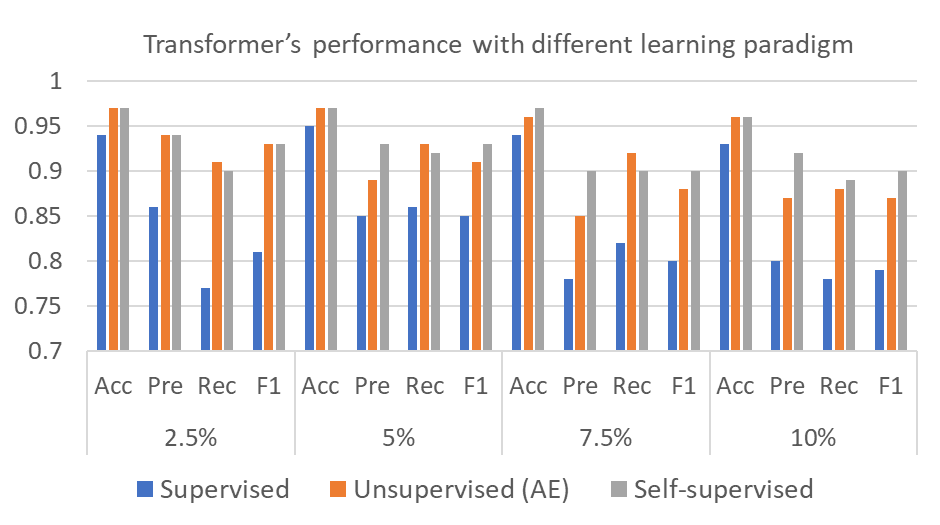}
	\caption{Transformer's performance with different learning paradigms including supervised, unsupervised (AE), and self-supervised contrastive learning (SmoothInfoNCE).}
	\label{fig:transformer_learningparadigm}
\end{figure*}

Furthermore, motivated by adapting the unsupervised learning from Azar 
 et al. \cite{azar2021deep}, and our previous study \cite{le2023adaptation}, we continue to experiment with Transformer for different learning paradigms, including supervised, unsupervised, and self-supervised contrastive learning (with Smooth InfoNCE loss). Table \ref{tab:transformer_learningparadigm}, and Fig. \ref{fig:transformer_learningparadigm} present a comparative analysis of a Transformer model's performance across three learning paradigms: supervised, unsupervised (using Autoencoders, AE), and self-supervised. Across all metrics, self-supervised learning exhibits consistently high performance, particularly excelling over supervised learning. Even with a minimal 2.5\% of annotated data, self-supervised learning matches the top accuracy and precision shown by unsupervised learning while significantly surpassing supervised learning, especially in recall and F1 scores. As the amount of annotated data increases, the self-supervised Transformer maintains its superiority over supervised learning and demonstrates marked improvements over the unsupervised AE approach. This trend is consistent with up to 10\% annotated data, where self-supervised learning still shows the highest metrics, confirming its effectiveness in leveraging both labeled and unlabeled data to improve performance, particularly in scenarios with limited annotations.

\begin{table*}[]
\caption{A comparison of Transformer's performance with different learning paradigm}
\label{tab:transformer_learningparadigm}
\begin{tabular}{|l||llll||llll||llll||llll|}
\hline
\multicolumn{1}{|c|}{\multirow{2}{*}{Transformer}} &
  \multicolumn{4}{c||}{2.5\%} &
  \multicolumn{4}{c||}{5\%} &
  \multicolumn{4}{c||}{7.5\%} &
  \multicolumn{4}{c|}{10\%} \\ \cline{2-17} 
\multicolumn{1}{|c|}{} &
  \multicolumn{1}{l|}{Acc} &
  \multicolumn{1}{l|}{Pre} &
  \multicolumn{1}{l|}{Rec} &
  F1 &
  \multicolumn{1}{l|}{Acc} &
  \multicolumn{1}{l|}{Pre} &
  \multicolumn{1}{l|}{Rec} &
  F1 &
  \multicolumn{1}{l|}{Acc} &
  \multicolumn{1}{l|}{Pre} &
  \multicolumn{1}{l|}{Rec} &
  F1 &
  \multicolumn{1}{l|}{Acc} &
  \multicolumn{1}{l|}{Pre} &
  \multicolumn{1}{l|}{Rec} &
  F1 \\ \hline
Supervised &
  \multicolumn{1}{l|}{0.94} &
  \multicolumn{1}{l|}{0.86} &
  \multicolumn{1}{l|}{0.77} &
  0.81 &
  \multicolumn{1}{l|}{0.95} &
  \multicolumn{1}{l|}{0.85} &
  \multicolumn{1}{l|}{0.86} &
  0.85 &
  \multicolumn{1}{l|}{0.94} &
  \multicolumn{1}{l|}{0.78} &
  \multicolumn{1}{l|}{0.82} &
  0.80 &
  \multicolumn{1}{l|}{0.93} &
  \multicolumn{1}{l|}{0.80} &
  \multicolumn{1}{l|}{0.78} &
  0.79 \\ \hline
Unsupervised   (AE) &
  \multicolumn{1}{l|}{\textbf{0.97}} &
  \multicolumn{1}{l|}{\textbf{0.94}} &
  \multicolumn{1}{l|}{\textbf{0.91}} &
  \textbf{0.93} &
  \multicolumn{1}{l|}{\textbf{0.97}} &
  \multicolumn{1}{l|}{0.89} &
  \multicolumn{1}{l|}{\textbf{0.93}} &
  0.91 &
  \multicolumn{1}{l|}{0.96} &
  \multicolumn{1}{l|}{0.85} &
  \multicolumn{1}{l|}{\textbf{0.92}} &
  0.88 &
  \multicolumn{1}{l|}{\textbf{0.96}} &
  \multicolumn{1}{l|}{0.87} &
  \multicolumn{1}{l|}{0.88} &
  0.87 \\ \hline
Self-supervised &
  \multicolumn{1}{l|}{\textbf{0.97}} &
  \multicolumn{1}{l|}{\textbf{0.94}} &
  \multicolumn{1}{l|}{0.90} &
  \textbf{0.93} &
  \multicolumn{1}{l|}{\textbf{0.97}} &
  \multicolumn{1}{l|}{\textbf{0.93}} &
  \multicolumn{1}{l|}{0.92} &
  \textbf{0.93} &
  \multicolumn{1}{l|}{\textbf{0.97}} &
  \multicolumn{1}{l|}{\textbf{0.90}} &
  \multicolumn{1}{l|}{0.90} &
  \textbf{0.90} &
  \multicolumn{1}{l|}{\textbf{0.96}} &
  \multicolumn{1}{l|}{\textbf{0.92}} &
  \multicolumn{1}{l|}{\textbf{0.89}} &
  \textbf{0.90} \\ \hline
\end{tabular}
\end{table*}

\section{Conclusion}
\label{sec:conclusion}
In the context of PPG signal analysis for PICU applications, this study presents SSL as a promising approach to utilizing the vast amount of unlabeled data. By integrating SSL into Transformer models, the results demonstrated improved capacity for signal interpretation and enhanced accuracy in detecting PPG artifacts. Among the various SSL techniques explored—masking, contrastive learning, and DINO—contrastive learning emerged as the most effective, particularly in scenarios with limited labeled data.\\

The research proposed refinements to contrastive learning to optimize SSL further, focusing on modifying the loss functions inspired by the original InfoNCE loss. These modifications enabled more stable training and better convergence, resulting in improved performance of the Transformer for artifact identification tasks.

However, this study is limited by its exclusive focus on PPG data collected within a single PICU environment, potentially restricting the generalizability of the findings. Future research should extend this work to more extensive and diverse benchmarking datasets \cite{liang2018new, maqsood2021benchmark} to comprehensively evaluate the efficacy of SSL techniques in Transformer models for PPG analysis. Such validation will help establish the robustness and applicability of SSL approaches across varied clinical settings and datasets. Additionally, while our current approach using Transformer-based SSL achieves high accuracy in detecting PPG signal artifacts, it does not provide interpretability in its decision-making process. Integrating attention map analysis could help uncover which signal regions or patterns contribute most to artifact detection, thereby enhancing model transparency. Future work will explore this approach to improve interpretability, enabling a clearer understanding of the model’s decisions for clinical and practical applications.

This research underscores SSL's transformative potential in elevating neural networks' performance using limited labeled data, offering a viable path for future advancements in data-scarce environments like the PICU.

\section*{Acknowledgment}

This work was supported in part by the Natural Sciences and Engineering Research Council (NSERC), in part by the Institut de Valorisation des données de l’Université de Montréal (IVADO), in part by the Fonds de la recherche en sante du Quebec (FRQS).

\bibliographystyle{IEEEtran}
\bibliography{IEEEabrv,Bibliography}



\begin{IEEEbiography}[{\includegraphics[width=1in, height=1.25in, clip, keepaspectratio]{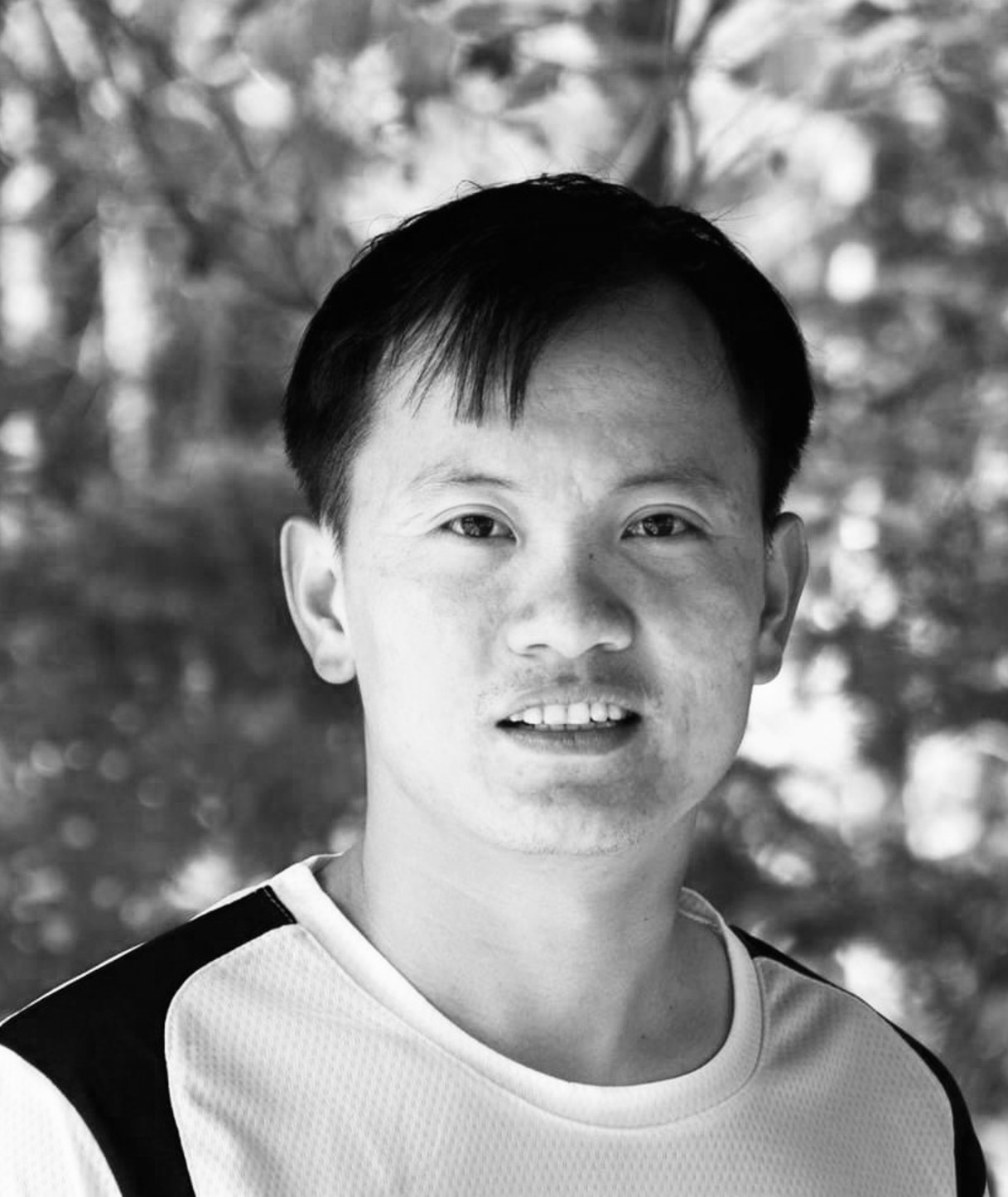}}]{Thanh-Dung Le} (Member, IEEE) received a B.Eng. degree in mechatronics engineering from Can Tho University, Vietnam, an M.Eng. degree in electrical engineering from Jeju National University, S. Korea, and a Ph.D. in electrical engineering (Major in Applied Artificial Intelligence) from Ecole de Technologie Superieure (ETS), University of Quebec, Canada. From October 2023 to May 2024, he was a Postdoctoral Fellow with the Biomedical Information Processing Laboratory, ETS. His research interests include applied machine learning approaches for biomedical informatics problems. Before that, he joined the Institut National de la Recherche Scientifique, Canada, where he researched classification theory and machine learning. He is currently a Research Associate at the Interdisciplinary Center for Security, Reliability, and Trust (SnT) at the University of Luxembourg, focusing on applied machine learning approaches for satellite communications systems. He received the merit doctoral scholarship from Le Fonds de Recherche du Quebec Nature et Technologies. He also received the NSERC-PERSWADE fellowship in Canada and a graduate scholarship from the Korean National Research Foundation, S. Korea.
\end{IEEEbiography}

\begin{IEEEbiography}[{\includegraphics[width=1in, height=1.25in, clip, keepaspectratio]{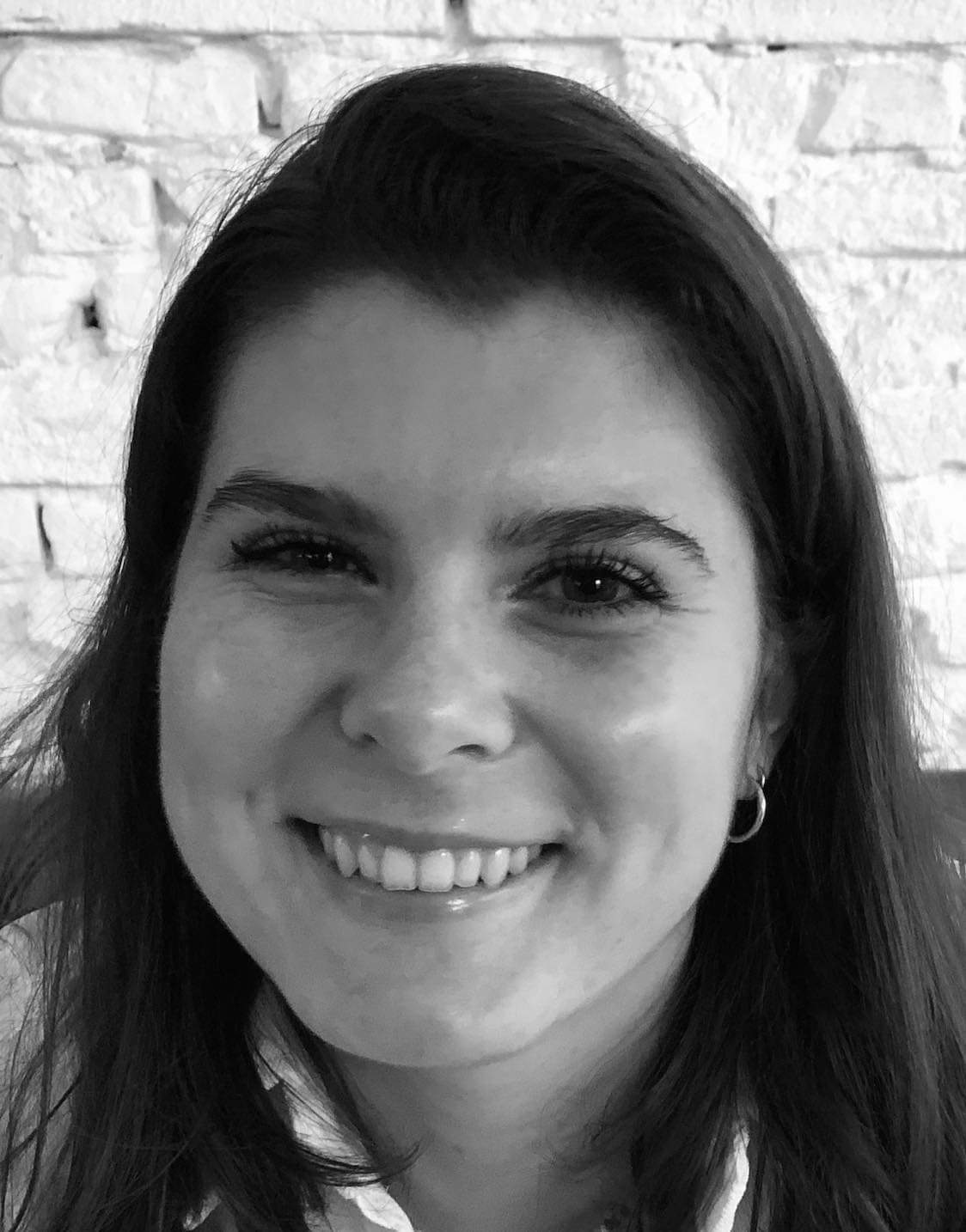}}]{Clara Macabiau} is a double degree student in Canada. After three years at the  \'{E}cole nationale sup\'{e}rieure d'\'{e}lectrotechnique, d'\'{e}lectronique, d'informatique, d'hydraulique et des t\'{e}l\'{e}communications (ENSEEIHT) engineering school in Toulouse, she is completing her master's degree in electrical engineering at \'{E}cole de Technologie Sup\'{e}rieure (ETS), Canada. Her master's project focuses on the detection of artifacts in photoplethysmography signals. She interests in signal processing, machine learning, and electronics.
\end{IEEEbiography}

\begin{IEEEbiography}[{\includegraphics[width=1in, height=1.25in, clip, keepaspectratio]{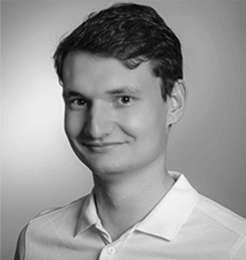}}]{Kevin Albert} is a physiotherapist who graduated from EUSES School of Health and Sport (2018 - Girona, Spain). He developed clinical expertise in the field of function rehabilitation after neuro-traumatic injury (France) and in cardio-respiratory rehabilitation (Swiss). He is currently enrolled in the Master's Biomedical Engineering program at the University of Montreal and has joined the Clinical Decision Support System (CDSS) laboratory under the supervision of Prof. P. Jouvet, M.D. Ph.D. in the Pediatric Intensive Care Unit at Sainte-Justine Hospital (Montréal, Canada) since May 2023. His primary research interest is the application of new technologies of support care system tools with artificial intelligence, especially in ventilatory support. His research program is supported by the Sainte-Justine Hospital and the Quebec Respiratory Health Research Network (QRHN).
\end{IEEEbiography}

\begin{IEEEbiography}
[{\includegraphics[width=1in,height=1.25in, clip, keepaspectratio]{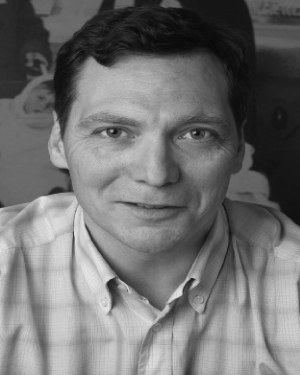}}]{Philippe Jouvet}received the M.D. degree from Paris V University, Paris, France, in 1989, the M.D. specialty in pediatrics and the M.D. subspecialty in intensive care from Paris V University, in 1989 and 1990, respectively, and the Ph.D. degree in pathophysiology of human nutrition and metabolism from Paris VII University, Paris, in 2001. He joined the Pediatric Intensive Care Unit of Sainte Justine Hospital—University of Montreal, Montreal, QC, Canada, in 2004. He is currently the Deputy Director of the Research Center and the Scientific Director of the Health Technology Assessment Unit, Sainte Justine Hospital–University of Montreal. He has a salary award for research from the Quebec Public Research Agency (FRQS). He currently conducts a research program on computerized decision support systems for health providers. His research program is supported by several grants from the Sainte-Justine Hospital, Quebec Ministry of Health, the FRQS, the Canadian Institutes of Health Research (CIHR), and the Natural Sciences and Engineering Research Council (NSERC). He has published more than 160 articles in peer-reviewed journals. Dr. Jouvet gave more than 120 lectures in national and international congresses.
\end{IEEEbiography}

\begin{IEEEbiography}[{\includegraphics[width=1in,height=1.25in, clip, keepaspectratio]{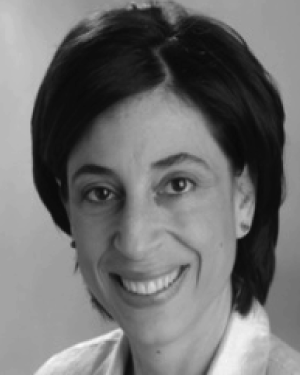}}]{Rita Noumeir} (Member, IEEE) received master's and Ph.D. degrees in biomedical engineering from École Polytechnique of Montreal. She is currently a Full Professor with the Department of Electrical Engineering, École de Technologie Superieure (ETS), Montreal. Her main research interest is in applying artificial intelligence methods to create decision support systems. She has extensively worked in healthcare information technology and image processing. She has also provided consulting services in large-scale software architecture, healthcare interoperability, workflow analysis, and technology assessment for several international software and medical companies, including Canada Health Infoway.
\end{IEEEbiography}

\EOD

\end{document}